%% file: acl_latex.tex
\documentclass[11pt]{article}

\usepackage[preprint]{acl}

\usepackage{times}
\usepackage{latexsym}
\usepackage{pifont} 
\usepackage[dvipsnames,table]{xcolor} 
\usepackage{graphicx} 

\usepackage[T1]{fontenc}

\usepackage[utf8]{inputenc}

\usepackage{microtype}

\usepackage{inconsolata}

\usepackage{graphicx}

\usepackage{amsmath}
\usepackage{booktabs}
\usepackage{tabularx}
\usepackage{tikz}
\usepackage{pgfplots}
\usepackage{placeins}
\pgfplotsset{compat=1.18}
\makeatletter
\IfFileExists{latex/sections/00_abstract.tex}{%
}{%
}
\makeatother
\newcommand{\proposedname}{\textsc{WorldReasoner}}

\title{\proposedname: Evaluating Whether Language Model Agents Forecast Events with Valid Reasoning}

\author{Yizhou Chi \quad Eric Chamoun \quad Zifeng Ding \quad Andreas Vlachos \\
  Department of Computer Science and Technology \\
  University of Cambridge \\
  \texttt{\{yc697, ec806, zd320, av308\}@cam.ac.uk} \\}

\begin{document}
\maketitle

\input{\sectionpath dataset_stats.tex}
\input{\sectionpath 00_abstract.tex}
\input{\sectionpath 01_introduction.tex}
\input{\sectionpath 02_related_work.tex}
\input{\sectionpath 03_benchmark.tex}
\input{\sectionpath 04_worldreasoner_pipeline.tex}

\input{\sectionpath 05_experiments.tex}

\input{\sectionpath 06_discussion.tex}
\input{\sectionpath 07_conclusion.tex}
\input{\sectionpath 09_limitations.tex}
\input{\sectionpath 10_ethics.tex}

\bibliography{custom}

\clearpage
\twocolumn
\appendix
\section*{Appendix}
\input{\sectionpath 99_appendix.tex}

\end{document}

%% file: sections/dataset_stats.tex

\newcommand{\wrResolvedQuestions}{345}
\newcommand{\wrDomains}{10}
\newcommand{\wrPolymarketQuestions}{97}
\newcommand{\wrNewsQuestions}{248}
\newcommand{\wrCollectedArticles}{14,141}
\newcommand{\wrExtractedEvents}{8,087}
\newcommand{\wrGraphQuestions}{345}
\newcommand{\wrAnnotationQuestions}{120}

\newcommand{\wrBinaryQuestions}{238}
\newcommand{\wrBinaryShare}{69.0\%}
\newcommand{\wrMCQQuestions}{46}
\newcommand{\wrMCQShare}{13.3\%}
\newcommand{\wrQuantityQuestions}{40}
\newcommand{\wrQuantityShare}{11.6\%}
\newcommand{\wrTimeframeQuestions}{21}
\newcommand{\wrTimeframeShare}{6.1\%}

\newcommand{\wrPoliticsQuestions}{88}
\newcommand{\wrCultureQuestions}{58}
\newcommand{\wrHealthQuestions}{54}
\newcommand{\wrSportsQuestions}{39}
\newcommand{\wrFinanceQuestions}{37}
\newcommand{\wrClimateQuestions}{25}
\newcommand{\wrTechnologyQuestions}{20}
\newcommand{\wrGeneralQuestions}{11}
\newcommand{\wrBusinessQuestions}{10}
\newcommand{\wrScienceQuestions}{3}

\newcommand{\wrStartDateRange}{March 29, 2023 to February 28, 2026}
\newcommand{\wrResolutionDateRange}{October 13, 2023 to April 11, 2026}

%% file: sections/00_abstract.tex
\begin{abstract}
Forecasting real-world events requires language-model agents to reason under uncertainty from incomplete, time-bounded information. Yet evaluating whether agents genuinely forecast requires more than final-answer accuracy: a model may be correct by recalling memorized training facts, citing fabricated evidence, or producing an unsupported causal story. We present \proposedname\footnotemark, an evaluation framework for temporally valid event forecasting. Each task gives an agent a resolved forecasting question, a simulated forecast date, and access only to evidence available before that date; after resolution, the framework scores the submitted probability, cited evidence, and optional causal event graph. \proposedname\ reports three complementary axes: outcome quality against resolved answers, evidence quality over cited sources, and reasoning quality against post-resolution hindsight graphs. The benchmark is built by an agentic construction pipeline that generates forecasting questions, collects time-stamped evidence, and builds hindsight reference graphs at scale, yielding \wrResolvedQuestions{} resolved tasks derived from \wrCollectedArticles{} articles with graphs covering \wrExtractedEvents{} extracted events. Across six controlled agent settings, temporally valid retrieval is the strongest driver of outcome accuracy; causal graph construction improves key-event recovery; and correct graph-enabled forecasts are more strongly grounded in key events and relevant sources, yet agents still struggle to convert grounded evidence into calibrated probabilities.

\end{abstract}
\footnotetext{Code and benchmark artifacts are available at \href{https://github.com/cyzus/worldreasoner/}{github.com/cyzus/worldreasoner}.}

%% file: sections/01_introduction.tex

\section{Introduction}
\label{sec:introduction}

To forecast real-world events, large language model (LLM) agents must reason from partial evidence under strict temporal constraints. As agents increasingly use search tools, databases, and structured reasoning modules, evaluation must measure whether they can make prospective judgments from only the information available at forecast time, rather than merely know the answer after the fact.

A fundamental obstacle is temporal data leakage. Static resolved benchmarks such as Autocast \cite{zou2022forecastingfutureworldevents} are fast and reproducible, but quickly become contaminated as LLM training corpora advance. An agent backed by a model trained through 2025 that answers a question about a 2022 election is not forecasting; it is recalling history. This temporal leakage makes outcome accuracy alone an unreliable measure of predictive reasoning.

\begin{figure*}[t]
    \centering
    \includegraphics[width=\linewidth]{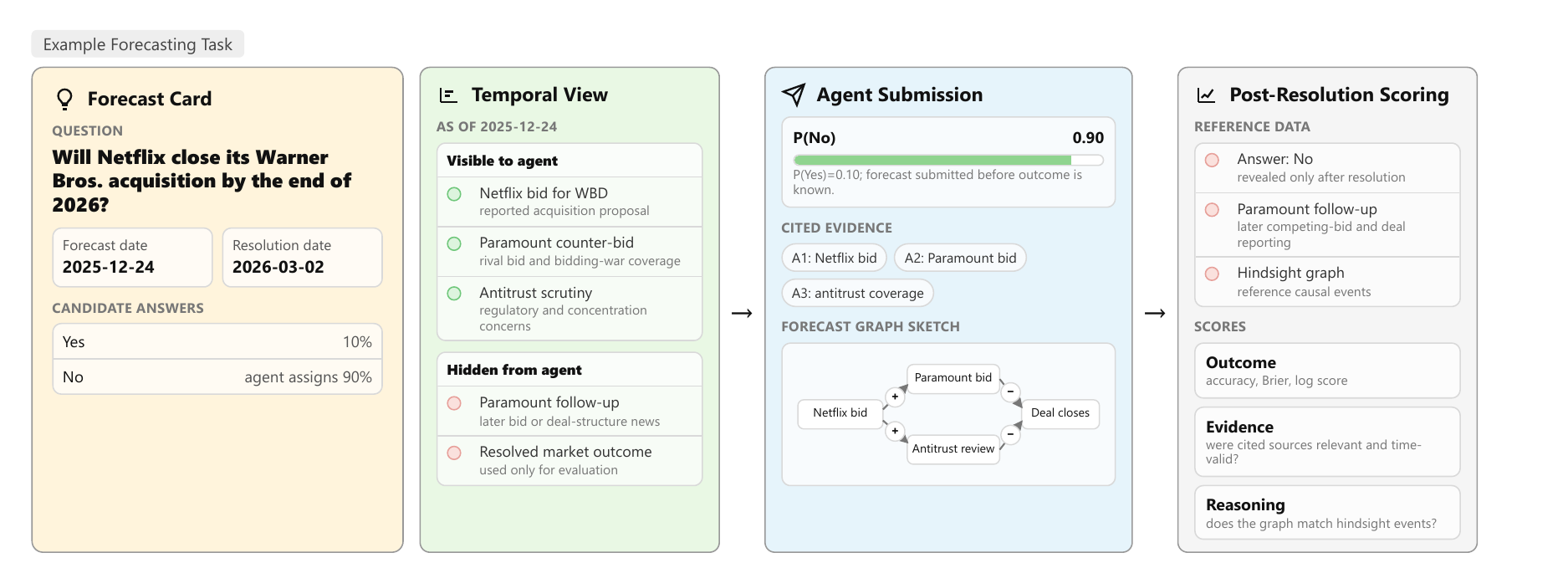}
    \caption{Example forecasting task in \proposedname. The agent answers a resolved question from a simulated forecast date, using only evidence visible at that time, and submits a probability, cited evidence, and a causal forecast graph. The graph shown is a simplified rendering of an actual submitted forecast graph for this task. After resolution, \proposedname\ evaluates outcome quality, evidence quality, and reasoning quality against held-out reference data.}
    \label{fig:task_example}
\end{figure*}

Live benchmarks address leakage by asking questions whose answers are still unknown. Systems such as RealTime QA \cite{kasai2024realtimeqawhatsanswer} and FutureX \cite{zeng2025futurexadvancedlivebenchmark} therefore guarantee a clean temporal boundary. However, live evaluation is inherently slow, since events may take weeks or months to resolve; it is difficult to scale; and it is hard to reproduce, because the information environment changes between evaluation runs. This creates a tradeoff between contamination control and experimental practicality.

Even with a clean temporal boundary, most benchmarks score only the resolved answer. A model may be correct by recalling memorized facts, citing hallucinated evidence, or producing an unsupported causal story; conversely, it may retrieve relevant evidence but fail to convert it into a calibrated probability. Forecasting is therefore also an evidence-grounding and causal-reasoning task.

We propose \proposedname, an evaluation framework for reproducible historical forecasting. Each task presents a resolved real-world question at a simulated historical forecast date and restricts all tool access to evidence available at that date. The agent submits a probabilistic forecast, sources, and optionally a causal event graph; \proposedname\ scores the submission immediately while preserving the temporal boundary that makes the task a forecast rather than a history lookup (Figure~\ref{fig:task_example}). A scalable construction pipeline generates questions from prediction markets and news streams and builds post-resolution hindsight graphs automatically.

\proposedname\ evaluates each submission along three axes: \textit{outcome quality}, whether the forecast assigns probability to the resolved answer; \textit{evidence quality}, whether cited or accessed sources are time-valid and relevant; and \textit{reasoning quality}, whether the forecast graph recovers hindsight-validated causal events. Together these axes distinguish outcome success from memorized recall, hallucinated grounding, and unsupported causal stories.

We evaluate representative frontier models across six controlled agent settings. Temporally valid retrieval is the strongest driver of outcome accuracy; explicit causal graph construction improves key-event recovery but does not by itself guarantee better answer selection. Conditional analysis shows that correct Search-Enabled Graph forecasts have higher key-event recall and source precision than incorrect forecasts, but this also exposes a remaining bottleneck: agents can retrieve relevant evidence and identify causal events, yet still fail to convert them into calibrated probabilities.

Our contributions are threefold:
\begin{enumerate}
    \item We introduce \proposedname, a temporally grounded forecasting evaluation framework that scores LLM agents along three axes: outcome quality, evidence quality, and reasoning quality.
    \item We propose an agentic benchmark construction pipeline that generates forecasting questions, collects time-stamped evidence, and builds post-resolution hindsight graphs, enabling scalable and reproducible expansion of the benchmark.
    \item We evaluate representative frontier models across controlled agent settings, showing that temporally valid retrieval drives outcome accuracy, evidence-grounded graph reasoning is associated with correctness, and converting grounded causal evidence into calibrated probabilities remains an open challenge.
\end{enumerate}

%% file: sections/02_related_work.tex

\section{Related Work}
\label{sec:related-work}

\begin{table*}[t]
\centering
\small
\setlength{\tabcolsep}{3.5pt}
\resizebox{\textwidth}{!}{%
\begin{tabular}{llcccc}
\toprule
\textbf{Paradigm} & \textbf{Benchmark} & \textbf{Immediate} & \textbf{Contam. Control} & \textbf{Reproducible} & \textbf{Reasoning Ref.} \\
\midrule
\textit{Static resolved} & Autocast \cite{zou2022forecastingfutureworldevents} & \textcolor{ForestGreen}{\ding{51}} & \textcolor{BrickRed}{\ding{55}} & \textcolor{ForestGreen}{\ding{51}} & \textcolor{BrickRed}{\ding{55}} \\
 & FOReCAst \cite{yuan2025forecastfutureoutcomereasoning} & \textcolor{ForestGreen}{\ding{51}} & \textcolor{BrickRed}{\ding{55}} & \textcolor{ForestGreen}{\ding{51}} & \textcolor{BrickRed}{\ding{55}} \\
\addlinespace[0.25em]
\textit{Live / continuous} & RealTime QA \cite{kasai2024realtimeqawhatsanswer} & \textcolor{BrickRed}{\ding{55}} & \textcolor{ForestGreen}{\ding{51}} & \textcolor{BrickRed}{\ding{55}} & \textcolor{BrickRed}{\ding{55}} \\
 & ForecastBench \cite{ICLR2025_ea74e45a} & \textcolor{BrickRed}{\ding{55}} & \textcolor{ForestGreen}{\ding{51}} & \textcolor{BurntOrange}{$\triangle$} & \textcolor{BrickRed}{\ding{55}} \\
 & MIRAI \cite{ye2024miraievaluatingllmagents} & \textcolor{ForestGreen}{\ding{51}} & \textcolor{BurntOrange}{$\triangle$} & \textcolor{ForestGreen}{\ding{51}} & \textcolor{BrickRed}{\ding{55}} \\
 & Daily Oracle \cite{dai2025llmsprescientcontinuousevaluation} & \textcolor{BrickRed}{\ding{55}} & \textcolor{ForestGreen}{\ding{51}} & \textcolor{BurntOrange}{$\triangle$} & \textcolor{BrickRed}{\ding{55}} \\
 & Prophet Arena \cite{yang2025llmasaprophetunderstandingpredictiveintelligence} & \textcolor{BrickRed}{\ding{55}} & \textcolor{ForestGreen}{\ding{51}} & \textcolor{BrickRed}{\ding{55}} & \textcolor{BurntOrange}{$\triangle$} \\
 & FutureX \cite{zeng2025futurexadvancedlivebenchmark} & \textcolor{BrickRed}{\ding{55}} & \textcolor{ForestGreen}{\ding{51}} & \textcolor{BrickRed}{\ding{55}} & \textcolor{BrickRed}{\ding{55}} \\
\addlinespace[0.25em]
\textit{Simulated-date replay} & Bench to the Future \cite{futuresearch2025benchtothefuture} & \textcolor{ForestGreen}{\ding{51}} & \textcolor{ForestGreen}{\ding{51}} & \textcolor{ForestGreen}{\ding{51}} & \textcolor{BrickRed}{\ding{55}} \\
\rowcolor{gray!10} & \textbf{\proposedname\ (Ours)} & \textbf{\textcolor{ForestGreen}{\ding{51}}} & \textbf{\textcolor{ForestGreen}{\ding{51}}} & \textbf{\textcolor{ForestGreen}{\ding{51}}} & \textbf{\textcolor{ForestGreen}{\ding{51}}} \\
\bottomrule
\end{tabular}%
}
\caption{Comparison of representative forecasting and time-sensitive evaluation benchmarks. \textcolor{ForestGreen}{\ding{51}} indicates explicit support, \textcolor{BrickRed}{\ding{55}} indicates no support, and \textcolor{BurntOrange}{$\triangle$} indicates partial support, e.g., the property holds only for some task types, evaluation stages, or replay settings. \proposedname\ combines immediate scoring, contamination control, reproducible simulated-date replay, and post-resolution reasoning references.}
\label{tab:benchmark_comparison}
\end{table*}

\paragraph{Forecasting benchmarks.}
Table~\ref{tab:benchmark_comparison} summarizes the benchmark landscape. Early LLM forecasting benchmarks such as Autocast \cite{zou2022forecastingfutureworldevents} and FOReCAst \cite{yuan2025forecastfutureoutcomereasoning} evaluate resolved world events, but fixed datasets become vulnerable to temporal leakage as model training corpora advance. Continuous and live benchmarks, including ForecastBench \cite{ICLR2025_ea74e45a}, RealTime QA \cite{kasai2024realtimeqawhatsanswer}, Daily Oracle \cite{dai2025llmsprescientcontinuousevaluation}, FutureX \cite{zeng2025futurexadvancedlivebenchmark}, Prophet Arena \cite{yang2025llmasaprophetunderstandingpredictiveintelligence}, and Mirai \cite{ye2024miraievaluatingllmagents}, improve temporal validity but are slow to resolve and hard to replay. Closest to our setting, Bench to the Future \cite{futuresearch2025benchtothefuture} uses simulated historical dates. \proposedname\ extends this replayable paradigm with automated task construction and post-resolution reasoning references.

\paragraph{Temporal and event-centered forecasting.}
Forecasting work spans numerical time-series settings and semantic event reasoning. ForecastTKGQA \cite{10.1007/978-3-031-47240-4_29} forecasts future events from temporal knowledge graphs, while Time-LLM \cite{jin2024timellmtimeseriesforecasting}, zero-shot language-model extrapolation \cite{gruver2024largelanguagemodelszeroshot}, TimeLlama \cite{yuan2023timellama}, and Time-R1 \cite{timer12025} study numerical or temporal reasoning. Work on temporal generalization further shows that models can exhibit temporal degradation and nostalgia bias \cite{zhu2025isyourllmoutdated}, motivating time-aware evaluation. \proposedname\ instead evaluates semantic event forecasts using dated articles, extracted events, and causal hypotheses as evidence and reasoning artifacts.

\paragraph{Reasoning verification and grounding.}
Our benchmark also relates to factuality, attribution, and causal event reasoning. AIS \cite{rashkin2022measuringattribution}, FActScore \cite{min-etal-2023-factscore}, FELM \cite{chen-etal-2023-felm}, and FACTS Grounding \cite{jacovi2025factsgrounding} evaluate evidence support, while Causal-BERT \cite{khetan2020causal}, ECHo \cite{xie2023echovisiolinguisticdatasetevent}, and CRAB \cite{romanou-etal-2023-crab} study event causality. \proposedname\ connects these threads to forecasting by evaluating answer accuracy, time-valid evidence, and alignment with post-resolution hindsight.

%% file: sections/03_benchmark.tex
\section{\proposedname}
\label{sec:benchmark}

\paragraph{Benchmark Usage.}
\proposedname\ is a controlled evaluation framework for historical event forecasting. For each resolved question, the framework supplies the question text, answer format, and a simulated forecast date. The agent submits an answer probability, cited or accessed evidence, and, when graph tools are enabled, a causal forecast graph. \proposedname\ then scores outcome, evidence, and reasoning quality.

\paragraph{Reproducible replay.}
The simulated-date design separates evaluation time from historical time. The evidence repository, question metadata, and resolution outcome are fixed for each task, while the simulated date is a controlled parameter chosen within the forecast window. The Temporal Gateway hides evidence and hindsight artifacts unavailable at that date, so later runs can reconstruct the same information state even after the event has resolved.

\begin{figure*}[t]
    \centering
    \includegraphics[width=\linewidth]{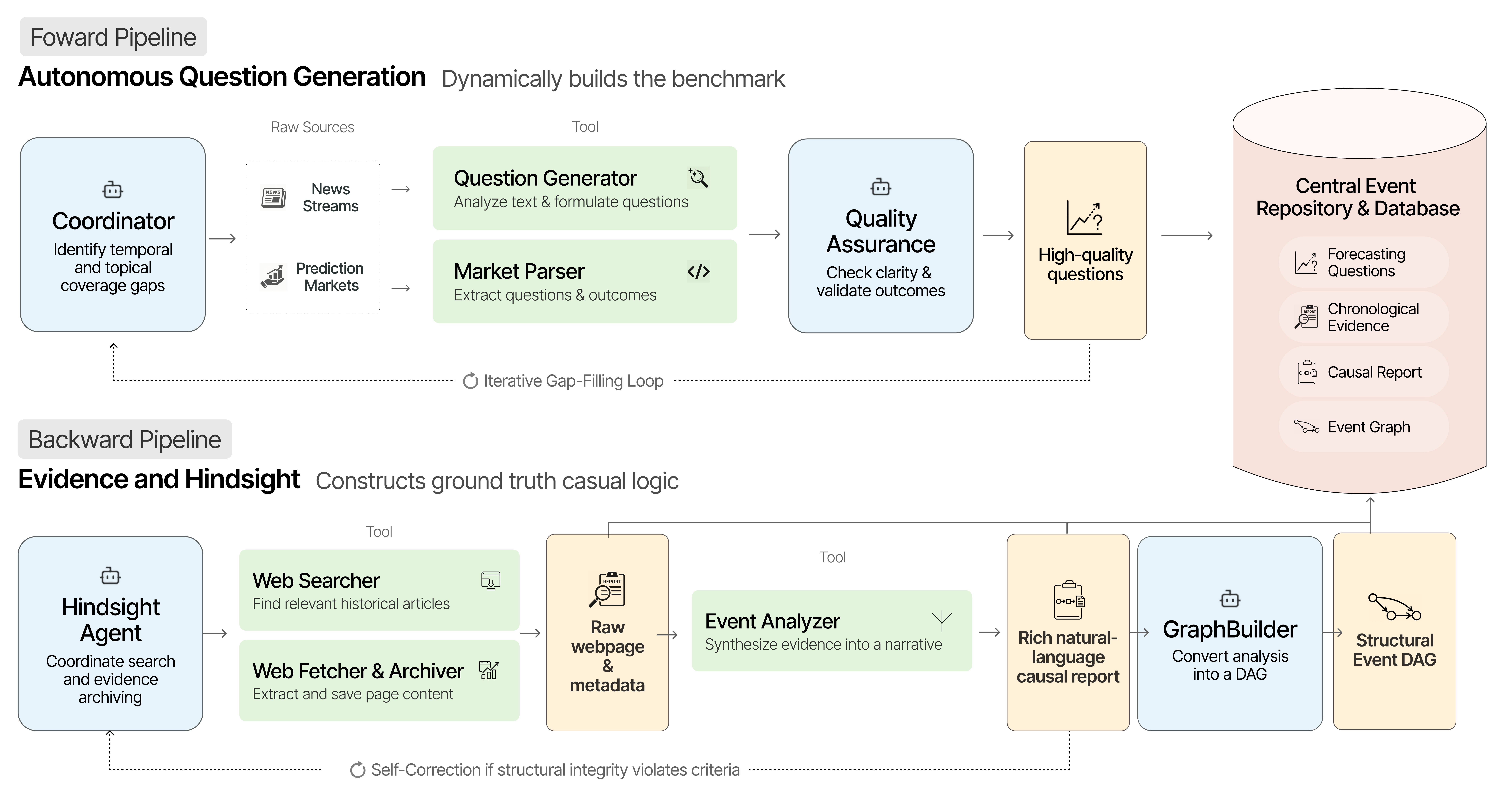}
    \caption{The forward and backward pipelines jointly construct the benchmark. The Forward Pipeline ingests news streams and prediction markets, generates forecasting questions, and filters them through quality assurance before storing them in the central event repository. The Backward Pipeline uses hindsight after resolution to collect evidence, synthesize causal explanations, and build structured event graphs used as reasoning references.}
    \label{fig:forward_backward_pipeline}
\end{figure*}

\subsection{Formal Task Definition}
Each benchmark instance is a resolved forecasting task
\begin{equation}
q_i = (x_i,\, t_i^{0},\, t_i^{\mathrm{sim}},\, t_i^{\mathrm{res}},\, y_i),
\end{equation}
where \(x_i\) is the question text, \(t_i^{0}\) is the earliest forecastable date, \(t_i^{\mathrm{sim}}\) is the simulated date given to the agent, \(t_i^{\mathrm{res}}\) is the resolution date, and \(y_i\) is the resolved outcome. Let \(\mathcal{A}_i\) denote the evidence repository associated with \(q_i\), and let \(\tau(a)\) be the timestamp of evidence item~\(a\). The Temporal Gateway exposes only the simulated-date evidence set
\begin{equation}
\mathcal{A}_i^{\mathrm{sim}} = \{a \in \mathcal{A}_i : \tau(a) < t_i^{\mathrm{sim}}\}.
\end{equation}
The agent produces a forecast
\begin{equation}
f_i = (\hat{y}_i,\, p_i,\, C_i,\, r_i,\, G_i^F),
\end{equation}
where \(\hat{y}_i\) is the predicted answer, \(p_i\) is the reported probability, \(C_i\) is the set of cited or accessed sources, \(r_i\) is the natural-language rationale, and \(G_i^F=(V_i^F,E_i^F)\) is the agent's forecast-time causal graph when graph tools are enabled. After resolution, the Hindsight Agent constructs a reference evidence set \(C_i^H\) and reference graph \(G_i^H=(V_i^H,E_i^H)\) from evidence available up to \(t_i^{\mathrm{res}}\).

\subsection{Three-Axis Evaluation}
Each task is evaluated along three complementary axes:
\begin{equation}
\begin{aligned}
\mathrm{Eval}(q_i) = \bigl(\,
  &S_{\mathrm{out}}(\hat{y}_i, y_i, p_i),\\
  &S_{\mathrm{evid}}(C_i, C_i^H),\\
  &S_{\mathrm{reason}}(G_i^F, G_i^H)\bigr),
\end{aligned}
\end{equation}
separating outcome quality, evidence quality, and reasoning quality rather than collapsing them into a single scalar.

\paragraph{Outcome quality} ($S_{\mathrm{out}}$) measures whether the forecast is correct: accuracy, Brier score, and log score.

\paragraph{Evidence quality} ($S_{\mathrm{evid}}$) measures whether the forecast drew on causally relevant sources, via source precision against the hindsight evidence corpus.

\paragraph{Reasoning quality} ($S_{\mathrm{reason}}$) measures whether the forecast graph recovers the events that proved causally important, via key-event recall against the hindsight graph. Key events are the top-$K$ hindsight events ranked by impact magnitude $\times$ confidence; a match requires both text similarity and date proximity.

\paragraph{Primary metrics.}
The main experiments report one headline metric for each axis. Outcome accuracy is
\begin{equation}
\mathrm{Acc} = \frac{1}{N}\sum_{i=1}^{N}\mathbf{1}[\hat{y}_i = y_i].
\end{equation}
For retrieval-enabled agents, source precision measures how many cited sources appear in the hindsight evidence set:
\begin{equation}
\mathrm{SrcP}_i = \frac{|C_i \cap C_i^H|}{|C_i|}.
\end{equation}
For graph-enabled agents, key-event recall measures recovery of the top-$K$ hindsight events \(K_i \subseteq V_i^H\):
\begin{equation}
\mathrm{KER}_i =
\frac{1}{|K_i|}
\sum_{e \in K_i}
\mathbf{1}\!\left[\exists\,\hat{e}\in V_i^F:\mathrm{match}(\hat{e},e)\right].
\end{equation}
We average SrcP and KER over applicable tasks; full definitions are in Appendix~\ref{app:metrics}.

\begin{figure*}[t]
\centering
\includegraphics[width=0.99\linewidth]{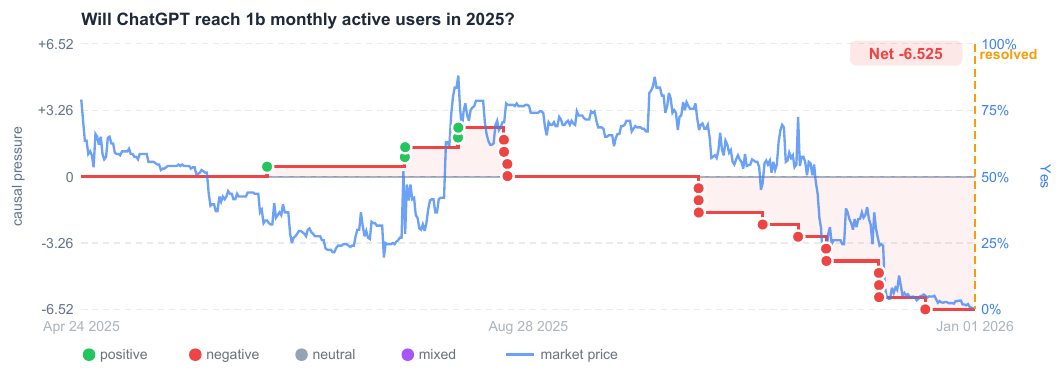}
\caption{From outcome label to reasoning reference. For ``Will ChatGPT reach 1 billion users?'' (No), each dot is a dated hindsight event linked to the outcome with a signed impact score; the step curve shows cumulative evidence direction over time; the blue line shows the contemporaneous Polymarket price. High-impact events define the reference set used by evidence and reasoning metrics.}
\label{fig:pressure_chatgpt}
\end{figure*}

%% file: sections/04_worldreasoner_pipeline.tex
\section{Benchmark Construction Pipeline}
\label{sec:system-architecture}

\proposedname\ is backed by two coordinated construction pipelines that create task instances and reference artifacts (Figure~\ref{fig:forward_backward_pipeline}).

\paragraph{Forward Pipeline.}
The forward pipeline ingests prediction markets and news streams. An LLM-driven generator proposes forecasting questions from article clusters, recording answer format, domain, resolution criteria, and provenance articles. For prediction markets, a parser maps market metadata into the same schema. A quality-assurance module scores questions for clarity, measurability, answerability, and temporal validity before they enter the benchmark database.

\paragraph{Backward Pipeline.}
After resolution, a \texttt{HindsightAgent} collects post-resolution evidence and synthesizes a causal explanation. A \texttt{GraphBuilderAgent} converts it into a structured event graph with dated event nodes and directed causal edges. Automated checks assess source diversity, temporal logic, and graph connectivity; a feedback loop triggers self-correction when quality criteria are not met. The resulting graphs serve as structural references for reasoning quality evaluation. The full schema and thresholds are in Appendix~\ref{app:hindsight_graph}.

Figure~\ref{fig:pressure_chatgpt} illustrates how a resolved question is converted into dated hindsight events with signed causal impact links.

\subsection{Hindsight Reference Quality}
Because hindsight graphs serve as references, we audit them separately from model forecasts. Human annotators review each hindsight event for factual accuracy, temporal correctness, source support, causal relevance, and uniqueness, using the protocol in Appendix~\ref{app:annotation}.

The reference graphs are not treated as perfect causal ground truth. Event nodes are checked for factual and temporal validity, while source overlap and key-event recall provide conservative evidence and reasoning signals. For Polymarket questions, hindsight pressure curves are also compared with market-price movements as an external sanity check (Appendix~\ref{app:pressure_charts}). Causal edge strengths and outcome-impact scores remain approximate, so reasoning metrics should be read as structured alignment with hindsight rather than exhaustive proof of causal validity.

\paragraph{Temporal Gateway.}
The Temporal Gateway enforces the simulated-date information boundary at query time. When a forecasting agent queries the article or event database, the gateway filters results to exclude any record dated after the simulated forecast date. Agents therefore operate on a pre-collected, time-bounded repository rather than live web access, making evaluation states reproducible. Per-model knowledge cutoff metadata is recorded separately and used to construct the contamination filter described in Section~\ref{sec:benchmark} (\S\,Handling Knowledge Cutoffs). Table~\ref{tab:dataset_statistics} summarizes the resulting benchmark scale.

\begin{table}[t]
\centering
\begin{tabular}{lr}
\toprule
\textbf{Statistic} & \textbf{Count} \\
\midrule
Resolved questions & \wrResolvedQuestions{} \\
Domains & \wrDomains{} \\
Polymarket questions & \wrPolymarketQuestions{} \\
News-derived questions & \wrNewsQuestions{} \\
Collected articles & \wrCollectedArticles{} \\
Extracted events & \wrExtractedEvents{} \\
Questions with hindsight graphs & \wrGraphQuestions{} \\
Annotation questions & \wrAnnotationQuestions{} \\
\bottomrule
\end{tabular}
\caption{Summary statistics for the \proposedname\ benchmark. Detailed answer-format, domain, and temporal distributions are reported in Appendix~\ref{app:dataset_statistics}.}
\label{tab:dataset_statistics}
\end{table}

%% file: sections/05_experiments.tex

\begin{figure*}[t]
\centering
\includegraphics[width=0.72\textwidth]{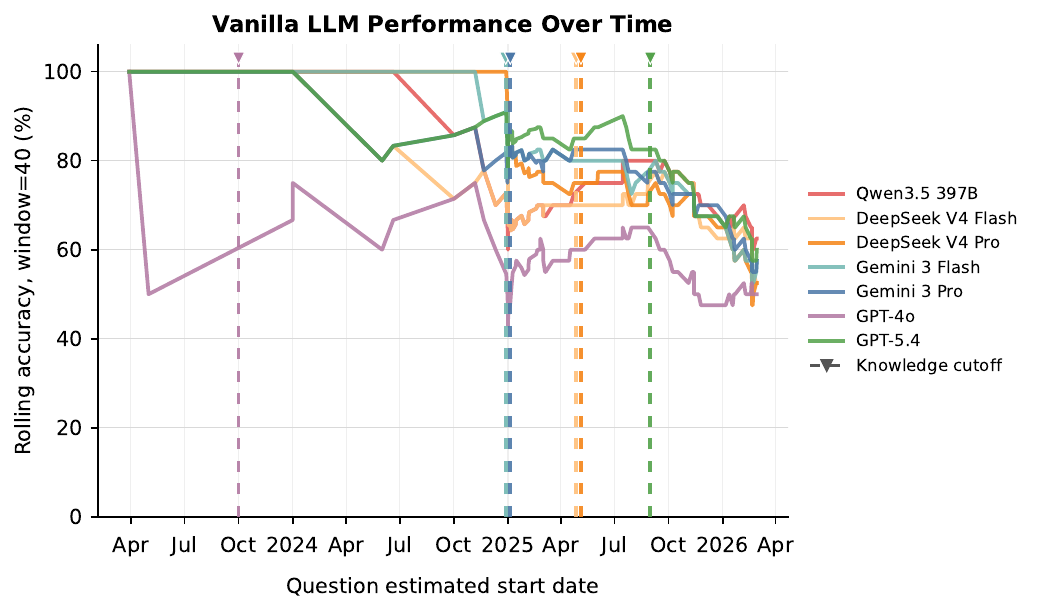}
\caption{Vanilla-only rolling accuracy by question start time. Since Vanilla LLM uses no retrieval or tools, post-cutoff declines motivate contamination-filtered reporting.}
\label{fig:vanilla_time_performance_main}
\end{figure*}

\section{Experiments}
\label{sec:evaluation}
We evaluate forecasting performance across six agent settings that progressively expand the information and reasoning available at a simulated historical date. In temporally constrained settings, the Temporal Gateway exposes only evidence available at the simulated forecast time. 

\subsection{Evaluation Setup}
\label{sec:setup}

\paragraph{Agent conditions.}
Six conditions expand agent capabilities: \textit{Vanilla LLM} uses parametric knowledge only; \textit{Causal Simulation} adds internal causal graph reasoning; \textit{Search-Enabled} adds simulated-date retrieval; \textit{Search-Enabled Graph} combines retrieval and graph construction; \textit{Near-Resolution Topline} sets the Temporal Gateway one day before resolution; and \textit{Real-Time Agent} uses live web access as a non-reproducible upper-bound reference.

\paragraph{Simulated forecast date.}
Unless otherwise specified, the simulated forecast date is set to the midpoint of each question's forecast window, between the estimated start date and the resolution date. The temporal-sensitivity experiment in Section~\ref{sec:sliding_window} varies this date across Early, Mid, Late, Near-Resolution, and Real-Time settings.

\paragraph{Models and question subset.}
The main filtered comparison evaluates Gemini 3 Flash/Pro, DeepSeek V4 Flash/Pro, GPT-4o, and GPT-5.4. All conditions start from the same fixed 120-question stratified subset; contamination filtering is applied per model-question pair, producing different filtered sample sizes by model.

\paragraph{Contamination filtering.}
The contamination filter excludes a model-question pair when the question's estimated start time precedes the model's knowledge cutoff, as the model may have encountered resolution-relevant information during pre-training. The Temporal Gateway controls run-time retrieval; the cutoff filter controls which pairs are included in cross-model comparison. Per-model cutoffs and filtering effects are in Appendix~\ref{app:contamination}.

\paragraph{Metrics.}
Each submission is evaluated along three axes (Section~\ref{sec:benchmark}; Appendix~\ref{app:metrics}): accuracy/Brier/log score for outcome quality, source precision (SrcP) for evidence quality when retrieval is available, and key-event recall (KER) for graph reasoning. KER is intentionally strict: it requires recovery of top-5 hindsight events under both textual similarity and a 7-day date gate, so absolute values are conservative.

\begin{table*}[t]
\centering
\small
\setlength{\tabcolsep}{3.0pt}
\begin{tabular}{@{}lc r rr rr rrr rrr rr@{}}
\toprule
& & \multicolumn{1}{c}{\textit{Vanilla}} & \multicolumn{2}{c}{\textit{Causal Sim.}} & \multicolumn{2}{c}{\textit{Search-En.}} & \multicolumn{3}{c}{\textit{SE Graph}} & \multicolumn{3}{c}{\textit{Near-Res.}} & \multicolumn{2}{c}{\textit{Real-Time}} \\
\cmidrule(lr){3-3}\cmidrule(lr){4-5}\cmidrule(lr){6-7}\cmidrule(lr){8-10}\cmidrule(lr){11-13}\cmidrule(lr){14-15}
\textbf{Model} & \textbf{Cutoff} & Acc & Acc & KER & Acc & SrcP & Acc & KER & SrcP & Acc & KER & SrcP & Acc & KER \\
\midrule
Gemini Flash      & Jan 2025 & 65.7 & \textbf{64.8} & 5.2 & 72.2 & 58.8 & 64.8 &  6.4 & 52.5 & 76.9 & 11.2 & 64.8 & 88.0 &  9.8 \\
Gemini Pro        & Jan 2025 & \textbf{66.7} & 59.3 & 5.2 & \textbf{74.1} & \textbf{70.4} & \textbf{73.1} & 12.2 & \textbf{66.6} & \textbf{79.6} & 13.3 & \textbf{71.6} & 91.7 & 10.4 \\
DS V4 Flash       & May 2025 & 52.9 & 57.4 & 4.7 & 73.5 & 65.2 & 69.1 & 12.0 & 57.4 & 77.9 & 18.1 & 66.6 & \textbf{95.6} & \textbf{15.0} \\
DS V4 Pro         & May 2025 & 54.4 & 57.4 & 5.7 & 67.6 & 65.2 & 66.2 & \textbf{13.7} & 58.7 & 76.5 & \textbf{20.4} & 71.4 & \textbf{95.6} & 12.0 \\
GPT-4o            & Oct 2023 & 53.4 & 46.6 & 3.4 & 59.3 & 59.7 & 55.1 &  7.2 & 56.0 & 66.9 & 12.6 & 70.4 & 82.2 &  8.2 \\
GPT-5.4           & Aug 2025 & 52.7 & 54.5 & \textbf{6.3} & 67.3 & 61.2 & 58.2 &  9.5 & 54.3 & 70.9 & 14.9 & 68.1 & 81.8 &  9.8 \\
\midrule
\textbf{Weighted avg.} & -- & 58.7 & 56.6 & 4.9 & 68.8 & 62.9 & 64.4 & 9.8 & 58.5 & 74.7 & 14.4 & 68.4 & 88.8 & 10.5 \\
\bottomrule
\end{tabular}
\caption{Per-model three-axis results across all six conditions (contamination-filtered). DS\,=\,DeepSeek. Acc\,=\,outcome quality (\%); KER\,=\,reasoning quality (\%, 7-day date gate); SrcP\,=\,evidence quality (\%); see Section~\ref{sec:setup} for definitions. Only metrics defined for each condition are shown: Vanilla has no graph or retrieval, Causal Simulation has no retrieval, Search-Enabled has no graph, and Real-Time lacks article provenance for SrcP. The weighted-average row averages over 525 contamination-filtered model-question pairs. Brier and log scores in Appendix~\ref{app:model_results}. GPT-5.4's unfiltered Vanilla accuracy is 69.2\%.}
\label{tab:permodel_results}
\end{table*}

\subsection{Aggregate Results}
\label{sec:results}

We report contamination-filtered scores; Table~\ref{tab:permodel_results} gives the full per-model breakdown across all six conditions.

Temporally situated retrieval is the primary driver of accuracy: weighted over the filtered per-model rows in Table~\ref{tab:permodel_results}, Causal Simulation does not improve over Vanilla LLM (56.6\% vs.\ 58.7\%), while Search-Enabled reaches 68.8\%. Near-Resolution reaches 74.7\%, confirming corpus solvability close to resolution; Real-Time reaches 88.8\% but is not a reproducible historical setting. Search-Enabled Graph trails search-only accuracy (64.4\% vs.\ 68.8\%), but it is the first retrieval-based setting that produces forecast graphs, yielding 9.8\% KER. This suggests that graph construction adds measurable causal-event recovery, although it does not automatically improve answer selection.

\subsection{Temporal Sensitivity}
\label{sec:sliding_window}

\begin{figure}[t]
\centering
\includegraphics[width=\linewidth]{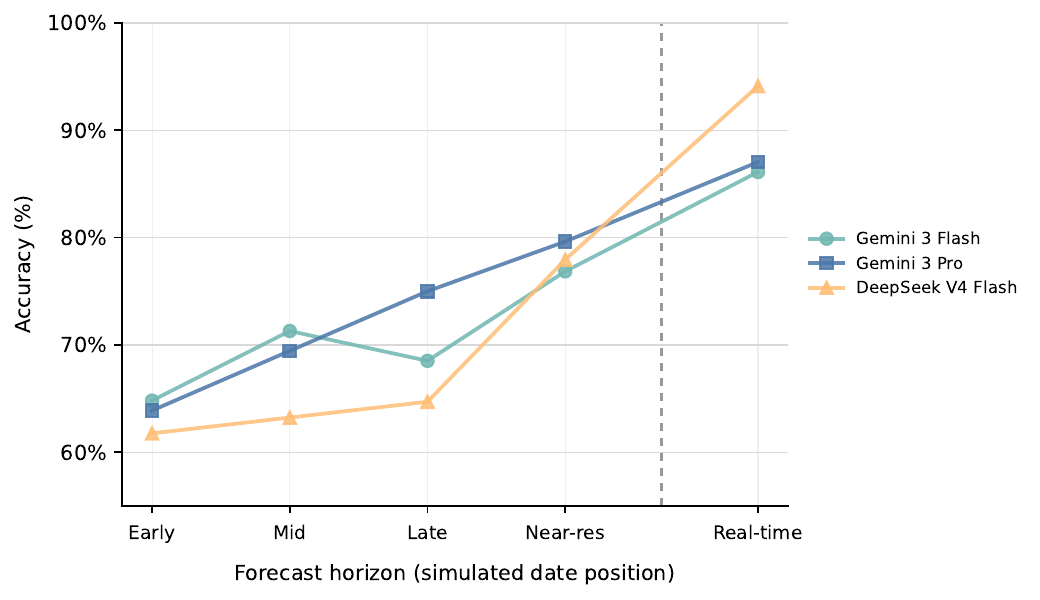}
\caption{Accuracy of the Search-Enabled Graph Agent across simulated forecast horizons. Accuracy generally rises as the horizon approaches resolution, confirming that the benchmark rewards later access to evidence.}
\label{fig:sliding_window}
\end{figure}

We run the Search-Enabled Graph Agent at five horizon positions for Gemini 3 Flash/Pro and DeepSeek V4 Flash. Early/Mid/Late correspond to the 0--33\%, 33--67\%, and 67--95\% percentiles of each question's forecast window; Near-res is set one day before resolution; Real-time uses live web access as an upper bound. Figure~\ref{fig:sliding_window} shows accuracy rising from Early ($\approx$62--65\%) through Near-res ($\approx$77--80\%) to Real-time ($\approx$86--94\%), confirming that the Temporal Gateway makes early forecasts genuinely harder.

\subsection{Per-Model Analysis}
\label{sec:per_model_analysis}

Table~\ref{tab:permodel_results} shows model heterogeneity: Gemini 3 Pro has the highest SE Graph accuracy and source precision, while DeepSeek V4 Pro has the highest SE Graph KER. KER improves from SE Graph to Near-Resolution across models, indicating that reasoning quality is gated by temporal evidence access.

\subsection{Correctness and Grounding}
\label{sec:conditional_quality}

\begin{table}[t]
\centering
\small
\setlength{\tabcolsep}{4pt}
\begin{tabular}{llrrr}
\toprule
\textbf{Condition} & & \textbf{N} & \textbf{KER} & \textbf{SrcP} \\
\midrule
Causal Sim.     & Correct   & 297 &  5.1 & --   \\
                & Incorrect & 228 &  4.7 & --   \\
\addlinespace[0.3em]
Search-Enabled  & Correct   & 361 & --   & 63.7 \\
                & Incorrect & 164 & --   & 61.5 \\
\addlinespace[0.3em]
SE Graph        & Correct   & 338 & 10.8 & 62.6 \\
                & Incorrect & 187 &  8.0 & 48.0 \\
\addlinespace[0.3em]
Near-Res.       & Correct   & 392 & 15.0 & 70.1 \\
                & Incorrect & 133 & 12.9 & 64.8 \\
\bottomrule
\end{tabular}
\caption{Key-event recall (KER, \%) and source precision (SrcP, \%) conditioned on forecast correctness (contamination-filtered). In Causal Simulation the near-zero KER gap shows that causal chain quality is only weakly correlated with outcome accuracy. SE Graph shows the largest SrcP gap, indicating that source quality is a stronger predictor of correctness when agents both retrieve and reason explicitly.}
\label{tab:conditional_quality}
\end{table}

Table~\ref{tab:conditional_quality} conditions grounding metrics on forecast correctness. Causal Simulation shows almost no KER gap (5.1\% vs.\ 4.7\%), and Search-Enabled shows only a small SrcP gap (63.7\% vs.\ 61.5\%). The separation is larger only when retrieval and graph construction are combined: correct SE Graph forecasts have higher KER (10.8\% vs.\ 8.0\%) and much higher SrcP (62.6\% vs.\ 48.0\%). Thus, grounding matters most when causal graphs are built from temporally valid evidence.

%% file: sections/06_discussion.tex
\section{Discussion}
\label{sec:discussion}

\paragraph{Three-axis evaluation catches what accuracy misses.}
Table~\ref{tab:conditional_quality} shows that correctness is only imperfectly coupled to grounding: Causal Simulation barely separates correct from incorrect forecasts, while SE Graph shows a larger but still incomplete separation. This is the central reason for evaluating outcome, evidence, and reasoning separately. A single-axis leaderboard cannot tell whether an agent was right because it retrieved useful time-valid evidence, because it reconstructed the relevant causal chain, or because the model had memorized post-resolution facts. Conversely, it cannot reveal cases where an agent retrieved useful evidence but failed at the final probability conversion step.

\paragraph{Per-model heterogeneity in correctness--grounding coupling.}
The aggregate gaps also mask model-level variation (Appendix~\ref{app:conditional_quality}). In SE Graph, Gemini 3 Pro and GPT-4o show large SrcP gaps between correct and incorrect forecasts, while DeepSeek V4 Flash shows the opposite sign: its incorrect forecasts sometimes cite more relevant sources than its correct ones. These splits are small after filtering and should not be read as definitive rankings, but they illustrate why the benchmark is useful as an analysis tool. Models with similar aggregate accuracy may differ in whether correctness is tied to evidence retrieval, event recovery, or downstream answer selection.

\paragraph{Why graph construction may not improve accuracy.}
The results also clarify why explicit graph construction can improve reasoning metrics without improving accuracy. The graph encourages agents to identify dated events and causal links, which raises KER, but the final answer still requires weighing conflicting evidence and converting qualitative causal pressure into a calibrated probability. GPT-4o illustrates this failure mode: despite the lowest SE Graph accuracy (55.1\%), it achieves 12.6\% Near-Resolution KER, on par with higher-accuracy models such as Gemini 3 Pro (13.3\%) and GPT-5.4 (14.9\%). This suggests that the next bottleneck is not only retrieval or graph construction, but the decision layer that maps grounded causal evidence to a calibrated forecast probability.

\paragraph{Contamination is the ``memorized facts'' failure mode.}
GPT-5.4's unfiltered Vanilla accuracy (69.2\%) looks competitive, but removing pairs whose start dates precede its August 2025 cutoff drops accuracy to 52.7\%. DeepSeek V4's May 2025 cutoff also excludes 52 pairs and reduces Vanilla accuracy by 7.1--10.6 points. Without the contamination filter, knowledge-only comparisons conflate memorization with forecasting ability. The Temporal Gateway and the cutoff filter therefore address different forms of temporal validity: the gateway constrains run-time evidence access, while the filter controls whether a model could already know resolution-relevant facts from pre-training.

%% file: sections/07_conclusion.tex
\section{Conclusion}
\label{sec:conclusion}

We presented \proposedname, an evaluation framework for LLM agent forecasting under strict temporal validity constraints. Agents are situated at a simulated historical date, with all evidence access mediated by a Temporal Gateway that enforces the information boundary reproducibly. Post-resolution hindsight graphs and human-annotated event validity labels support three-axis evaluation: outcome quality, evidence quality, and reasoning quality.

Our results show that temporally situated evidence access substantially improves outcome quality, while explicit graph construction improves reasoning alignment without automatically improving answer selection. A forecast can be correct without time-valid evidence, or incorrect despite retrieving causally relevant sources (Table~\ref{tab:conditional_quality}). Reporting all three axes surfaces where forecasting failures arise, a distinction that matters when reasoning validity is as important as outcome accuracy.

%% file: sections/09_limitations.tex
\section*{Limitations}

\paragraph{Coverage Bias}
News-based benchmarks reflect media coverage patterns, so less-covered domains are underrepresented and fast-moving public topics are over-sampled.

\paragraph{Resolution and Time Granularity}
Some questions have ambiguous resolutions in subjective or institution-dependent domains, and day-level Temporal Gateway access can miss intra-day leakage. Explicit criteria mitigate these issues, but some outcomes remain definition-sensitive.

\paragraph{Model Knowledge Cutoffs}
Training-data cutoffs vary across providers and are imperfect proxies for the information present in a model's parameters.

\paragraph{Evaluation and Annotation Scope}
Human annotation covers a fixed subset, so graph-quality metrics remain approximate. Hindsight graphs may inherit source or model biases; review validates event-node factuality, dates, source support, relevance, and uniqueness, but not every causal edge or impact score. Exact source matching may undercount relevant evidence from different articles.

%% file: sections/10_ethics.tex
\section*{Ethical Considerations}

\paragraph{Intended Use and Risks}
\proposedname\ is a research evaluation framework for studying whether forecasting agents use time-valid evidence and causally relevant reasoning, not a decision-support system for high-stakes forecasting. To reduce over-interpretation of benchmark scores as real-world predictive reliability, we report decomposed outcome, evidence, and reasoning metrics rather than a single leaderboard score.

\paragraph{Data and Privacy}
The benchmark uses public news articles, prediction-market metadata, generated forecasting questions, model forecasts, and derived event-graph annotations. These sources may mention public figures, organizations, or events, but the benchmark is not designed to collect private user data or infer sensitive attributes. Released artifacts will be for research use and follow applicable source terms and artifact licenses.

\paragraph{Human Annotation}
Human annotators were recruited through Prolific, compensated at approximately US\$20 per hour, and shown task instructions and informed-consent information. They reviewed factuality, dates, source support, relevance, and uniqueness of generated hindsight events, without predicting outcomes or providing sensitive personal information.

%% file: sections/99_appendix.tex
\input{sections/appendix/A_dataset_statistics}
\input{sections/appendix/B_hindsight_graph}
\input{sections/appendix/C_annotation}

\input{sections/appendix/D_pressure_charts}
\input{sections/appendix/E_contamination}
\input{sections/appendix/F_evaluation_metrics}

\input{sections/appendix/G_model_results}
\input{sections/appendix/H_frontend_inspection}

\input{sections/appendix/I_ai_use}
\input{sections/appendix/J_artifact_terms}

%% file: sections/appendix/A_dataset_statistics.tex
\section{Dataset Statistics}
\label{app:dataset_statistics}

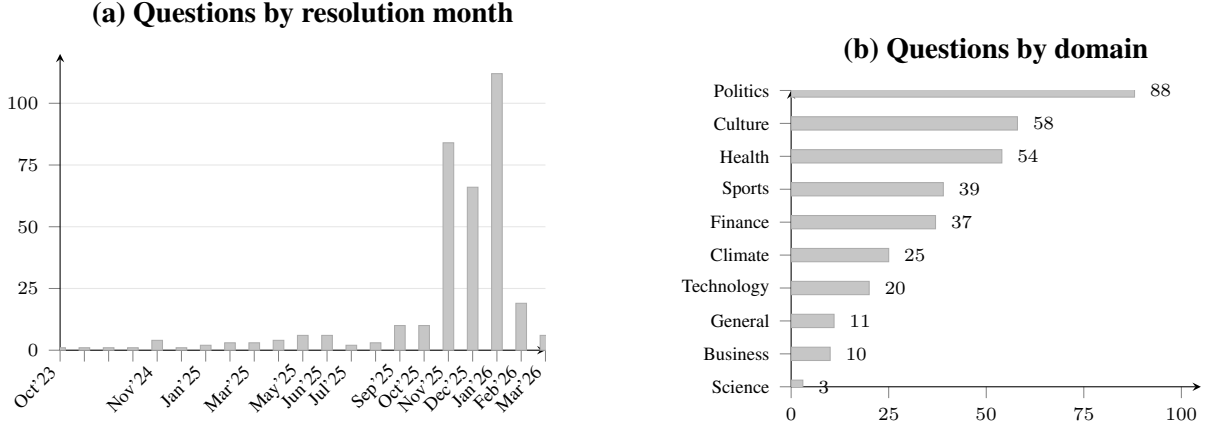
\begin{figure*}[h]
\centering
\begin{tikzpicture}
\begin{axis}[
  title={\textbf{(a) Questions by resolution month}},
  ybar,
  width=8.0cm, height=5.5cm,
  bar width=4pt,
  ymin=0, ymax=120,
  ytick={0,25,50,75,100},
  yticklabel style={font=\scriptsize},
  xtick=data,
  xticklabels={
    Oct'23,,,,Nov'24,,Jan'25,,Mar'25,,May'25,Jun'25,Jul'25,,Sep'25,Oct'25,Nov'25,Dec'25,Jan'26,Feb'26,Mar'26,Apr'26
  },
  xticklabel style={font=\scriptsize, rotate=45, anchor=east},
  enlarge x limits=0.04,
  axis x line=bottom,
  axis y line=left,
  tick align=outside,
  ymajorgrids=true,
  grid style={gray!20},
]
\addplot[fill=gray!45, draw=gray!65] coordinates {
  (1,  1)   
  (2,  1)   
  (3,  1)   
  (4,  1)   
  (5,  4)   
  (6,  1)   
  (7,  2)   
  (8,  3)   
  (9,  3)   
  (10, 4)   
  (11, 6)   
  (12, 6)   
  (13, 2)   
  (14, 3)   
  (15,10)   
  (16,10)   
  (17,84)   
  (18,66)   
  (19,112)  
  (20,19)   
  (21, 6)   
};
\end{axis}
\end{tikzpicture}%
\hfill%
\begin{tikzpicture}
\begin{axis}[
  title={\textbf{(b) Questions by domain}},
  xbar,
  width=7.0cm, height=5.5cm,
  bar width=5pt,
  xmin=0, xmax=105,
  xtick={0,25,50,75,100},
  xticklabel style={font=\scriptsize},
  ytick=data,
  yticklabels={Science, Business, General, Technology, Climate, Finance, Sports, Health, Culture, Politics},
  yticklabel style={font=\scriptsize},
  nodes near coords,
  nodes near coords style={font=\scriptsize, xshift=2pt},
  enlarge y limits=0.08,
  axis x line=bottom,
  axis y line=left,
  tick align=outside,
  ymajorgrids=false,
]
\addplot[fill=gray!45, draw=gray!65] coordinates {
  (3, 0)   
  (10,1)   
  (11,2)   
  (20,3)   
  (25,4)   
  (37,5)   
  (39,6)   
  (54,7)   
  (58,8)   
  (88,9)   
};
\end{axis}
\end{tikzpicture}
\caption{%
  \textbf{(a)} Resolution date distribution across the \wrResolvedQuestions{} resolved questions. The spike in late 2025 and early 2026 reflects the benchmark's coverage of that forecasting window.
  \textbf{(b)} Domain breakdown; Politics, Culture, and Health account for 58\% of questions.
  Estimated start dates range from \wrStartDateRange{}; resolution dates range from \wrResolutionDateRange{}.%
}
\label{fig:dataset_distribution}
\end{figure*}

The benchmark contains \wrResolvedQuestions{} resolved questions drawn from \wrCollectedArticles{} collected articles across \wrDomains{} domains. Questions originate from two sources: \wrNewsQuestions{} are derived from news articles and \wrPolymarketQuestions{} from Polymarket prediction markets. The dataset is dominated by binary questions (\wrBinaryShare{}) reflecting the prevalence of yes/no forecasting tasks in both sources, with a smaller number of multiple-choice, quantity, and timeframe questions covering more structured prediction targets. Across domains, Politics, Culture, and Health together account for 58\% of questions. Figure~\ref{fig:dataset_distribution} shows the resolution date distribution and domain breakdown; Table~\ref{tab:dataset_distribution} gives the corresponding counts.

The resolution date histogram (Figure~\ref{fig:dataset_distribution}a) shows that the majority of questions resolved in December 2025 through February 2026, with a long sparse tail extending back to late 2023. This reflects the pipeline's focus on a near-term forecasting window: questions are opened when events are uncertain and collected as they resolve, so the density naturally concentrates around recent resolution months. The sparse early tail consists of older Polymarket markets and long-running news events that resolved before the primary collection period.

\begin{table}[htbp]
\centering
\scriptsize
\setlength{\tabcolsep}{3pt}
\begin{tabular}{lrrlr}
\toprule
\textbf{Type} & \textbf{N} & \textbf{\%} & \textbf{Domain} & \textbf{N} \\
\midrule
Binary          & \wrBinaryQuestions{}   & \wrBinaryShare{}   & Politics   & \wrPoliticsQuestions{} \\
Multiple choice & \wrMCQQuestions{}      & \wrMCQShare{}      & Culture    & \wrCultureQuestions{}  \\
Quantity        & \wrQuantityQuestions{} & \wrQuantityShare{} & Health     & \wrHealthQuestions{}   \\
Timeframe       & \wrTimeframeQuestions{}& \wrTimeframeShare{}& Sports     & \wrSportsQuestions{}   \\
                &                        &                    & Finance    & \wrFinanceQuestions{}  \\
                &                        &                    & Climate    & \wrClimateQuestions{}  \\
                &                        &                    & Technology & \wrTechnologyQuestions{}\\
                &                        &                    & General    & \wrGeneralQuestions{}  \\
                &                        &                    & Business   & \wrBusinessQuestions{} \\
                &                        &                    & Science    & \wrScienceQuestions{}  \\
\bottomrule
\end{tabular}
\caption{Question type and domain distributions. Forecast horizons: $\leq$30\,d: 142; 31--90\,d: 32; $>$90\,d: 171; median: 86\,d; mean: 174\,d.}
\label{tab:dataset_distribution}
\end{table}

%% file: sections/appendix/B_hindsight_graph.tex
\section{Hindsight Graph Construction}
\label{app:hindsight_graph}

After a question resolves, the backward pipeline constructs a structured causal graph that serves as the reference artifact for both evidence and reasoning evaluation. This appendix describes the graph schema, the construction procedure, and the quality enforcement mechanism.

\subsection{Graph Schema}

A hindsight graph $G^H = (V^H, E^H, I^H)$ for question $q_i$ consists of event nodes $V^H$, causal edges $E^H$, and outcome impact records $I^H$.

\paragraph{Event nodes.}
Each node $v \in V^H$ represents a dated real-world event with the following fields:
\begin{itemize}  \item \textbf{Text}: a title and description of the event (validated for length and specificity).
  \item \textbf{Date}: the date the event occurred (\texttt{occurred\_date}), used for temporal matching during KER evaluation.
  \item \textbf{Type}: one of \textsc{Decision}, \textsc{Outcome}, \textsc{Indicator}, \textsc{Milestone}, or \textsc{External Shock}.
  \item \textbf{Source articles}: the set of article identifiers that document the event, used to compute source precision.
  \item \textbf{Review status}: \textsc{Pending}, \textsc{Approved}, \textsc{Rejected}, or \textsc{Revised}, set by human annotators (Appendix~\ref{app:annotation}).
\end{itemize}
One node is designated the outcome event, linked to the question's resolved answer.

\paragraph{Causal edges.}
Each directed edge $(u \to v) \in E^H$ represents a causal relationship between two events with:
\begin{itemize}  \item \textbf{Relation type}: one of \textsc{Causes}, \textsc{Enables}, \textsc{Prevents}, \textsc{Inhibits}, \textsc{Amplifies}, \textsc{Triggers}, or \textsc{Correlates}.
  \item \textbf{Strength} $s \in [-1, 1]$: causal effect magnitude, LLM-assigned with calibration rubric (0.1--0.3 weak; 0.4--0.6 moderate; 0.7--0.9 major).
  \item \textbf{Confidence} $c \in [0, 1]$: the graph builder's certainty in the link, based on available evidence.
  \item \textbf{Reasoning}: a natural-language mechanistic explanation.
  \item \textbf{Evidence articles}: supporting article identifiers.
\end{itemize}

\paragraph{Outcome impact records.}
Each record $\iota \in I^H$ captures a direct event-to-outcome impact:
\begin{itemize}  \item \textbf{Impact direction}: \textsc{Positive} (increases resolution likelihood), \textsc{Negative}, \textsc{Neutral}, or \textsc{Mixed}.
  \item \textbf{Impact magnitude} $m \in [0, 1]$: strength of the push/pull on the outcome (0.1--0.3 weak; 0.4--0.6 moderate; 0.7--0.9 major; 1.0 decisive). Neutral impacts are constrained to $m \leq 0.3$.
  \item \textbf{Confidence} $c \in [0, 1]$: certainty in the impact assessment.
\end{itemize}
The product $m \times c$ defines the event's \emph{impact score}, used to rank and select key events for KER evaluation.

\subsection{Construction Procedure}

The backward pipeline runs in two stages after a question resolves.

\paragraph{Stage 1: Evidence collection (HindsightAgent).}
The HindsightAgent retrieves post-resolution articles and synthesizes a causal explanation of how the outcome came about, identifying the key events and their temporal order.

\paragraph{Stage 2: Graph construction (GraphBuilderAgent).}
The GraphBuilderAgent converts the causal explanation into the structured graph schema. It uses dedicated tools to propose event nodes (\texttt{ProposeSubgraphTool}), record causal edges, and assign outcome impact records (\texttt{RecordOutcomeImpactTool}). All magnitude and confidence values are LLM-assigned under explicit calibration guidelines that require concrete evidence for high values; the guidelines discourage uniform scoring.

\subsection{Quality Enforcement}

After construction, the graph is validated against the following satisfaction thresholds:
\begin{itemize}  \item Minimum graph depth (causal chain length): $\geq 3$ hops from an initiating event to the outcome.
  \item Minimum event count: $\geq 10$ nodes.
  \item Minimum supporting articles: $\geq 20$ distinct sources.
  \item Minimum causal hypotheses: $\geq 1$ directed edge.
  \item Edge confidence $\geq 0.6$ and strength $\geq 0.3$ for included edges.
\end{itemize}
If any threshold is unmet, the GraphBuilderAgent automatically adds intermediate cause events and retries (\texttt{DeleteEventTool} / \texttt{DeleteHypothesisTool} are available for correction). This self-correction loop continues until all constraints are satisfied or the maximum iteration budget is exhausted. Graphs that fail to satisfy the thresholds are flagged (\texttt{graph\_built=False}) and excluded from evaluation.

\subsection{Key Event Selection for KER}

Key events $K_i \subseteq V^H$ are the top-$K$ nodes by impact score $m \times c$, where $K = 5$ in all reported experiments. Only nodes with an associated outcome impact record contribute to $K_i$; structural intermediate nodes without a direct impact assessment are excluded. This selection ensures that KER measures recovery of the most causally decisive events rather than all graph nodes.

%% file: sections/appendix/C_annotation.tex
\section{Annotation Protocol and Taxonomy}
\label{app:annotation}

Human annotators review hindsight graph events to validate the reference artifacts used for evidence and reasoning evaluation. Each event is assessed independently of any forecast agent's output. Annotators judge whether an event is: (1) real and factually accurate; (2) assigned the correct date; (3) supported by the cited source article; (4) causally relevant to the resolved outcome; and (5) not a duplicate of another event in the same graph. Events that pass all checks are approved; events that fail one or more checks are rejected with a reason label.

\begin{table}[h]
\centering
\scriptsize
\setlength{\tabcolsep}{3pt}
\begin{tabularx}{\columnwidth}{lrrX}
\toprule
\textbf{Quality signal} & \textbf{N} & \textbf{\%} & \textbf{Interpretation} \\
\midrule
Accepted events & 336/492 & 68.3 & Real, correctly dated, source-supported, relevant. \\
Factual-error rejections & 113/492 & 23.0 & Fabricated, wrong-date, or source-mismatch. \\
Skipped/unverifiable & 43/492 & 8.7 & Left unresolved rather than forced into a label. \\
\midrule
Source mismatch & 60/492 & 12.2 & Most common factual-error mode. \\
Wrong date & 27/492 & 5.5 & Event real but not at the graph date. \\
Fabricated & 26/492 & 5.3 & Could not be verified as described. \\
\bottomrule
\end{tabularx}
\caption{Annotation-quality summary. Annotation review serves as a conservative validation layer for hindsight graphs.}
\label{tab:annotation_quality}
\end{table}

\paragraph{Participant instructions and consent.}
Before annotation, participants were told that the task helps build causal event timelines for resolved forecasting questions. The instructions emphasized that annotators were not asked to predict the answer; instead, they were asked to verify whether each event card actually happened, was supported by the cited source, was dated correctly, and was causally relevant to the resolved question. Annotators were recruited through Prolific and paid approximately US\$20 per hour. The consent screen stated that participation was voluntary, that participants could withdraw at any time, that responses would be used for research purposes and handled confidentially, and that AI-assisted annotation was not permitted. The task then required annotators to review the question-level hindsight report before validating event cards chronologically.

\paragraph{Reject-reason taxonomy.}
\begin{itemize}
  \item \textbf{Fabricated}: The event did not occur as described, or no credible source confirms it.
  \item \textbf{WrongDate}: The event is real but is assigned an incorrect date (off by days, weeks, or more).
  \item \textbf{SourceMismatch}: The cited source article does not support the described event.
  \item \textbf{PredictionNotEvent}: The entry describes a forecast or prediction, not a historical occurrence.
  \item \textbf{Noise}: The event is real but causally irrelevant to the resolved outcome.
  \item \textbf{Duplicate}: The event is substantively identical to another event already in the graph.
  \item \textbf{TooBroad}: The event description is too vague or sweeping to be meaningfully evaluated.
  \item \textbf{Unverifiable}: The event cannot be confirmed or refuted from publicly available sources.
\end{itemize}

The taxonomy was developed from an initial pilot annotation session and revised before the main annotation phase. Event validity is assessed independently of the model-generated causal explanation: a real, temporally correct, source-supported event may be approved even if the attached impact reasoning is debatable. Table~\ref{tab:annotation_quality} summarizes the main quality checks for the annotation batch.

\paragraph{Annotation interface.}
The interface first presents task instructions and consent information, then requires annotators to review the resolved question and hindsight report before validating individual event cards.

%% file: sections/appendix/D_pressure_charts.tex
\section{Causal Pressure Charts}
\label{app:pressure_charts}

\paragraph{Chart construction.}
Pressure charts visualize hindsight reference graphs after resolution; they are not inputs to forecast agents. Dots are dated hindsight events, coloured by signed direction toward or against a ``Yes'' resolution. The step curve is cumulative causal pressure, $P_k = \sum_{i=1}^{k} d_i m_i c_i$, where direction $d_i \in \{+1,0,-1\}$ is weighted by impact magnitude $m_i$ and confidence $c_i$. For Polymarket questions, the blue line overlays the contemporaneous market price as an external check on the timing and direction of the hindsight graph. Annotation protocol and event-quality checks are in Appendix~\ref{app:annotation}.

\paragraph{Reference graph alignment with market moves.}
Using the reproducible market-check script, 81 Polymarket questions have enough pressure and price variation for a net-direction comparison; 75.3\% have the same sign for net cumulative pressure and net market movement. Among 83 questions with defined pressure/price correlations, 71.1\% are positive, with median correlation 0.595. A stricter event-local check gives 59.7\% agreement over 278 usable event-level checks, reflecting that markets often anticipate information before the event date or react to several signals at once.

\begin{figure*}[t]
\centering
\begin{minipage}{0.78\linewidth}
\centering
\includegraphics[width=\linewidth,height=0.20\textheight,keepaspectratio]{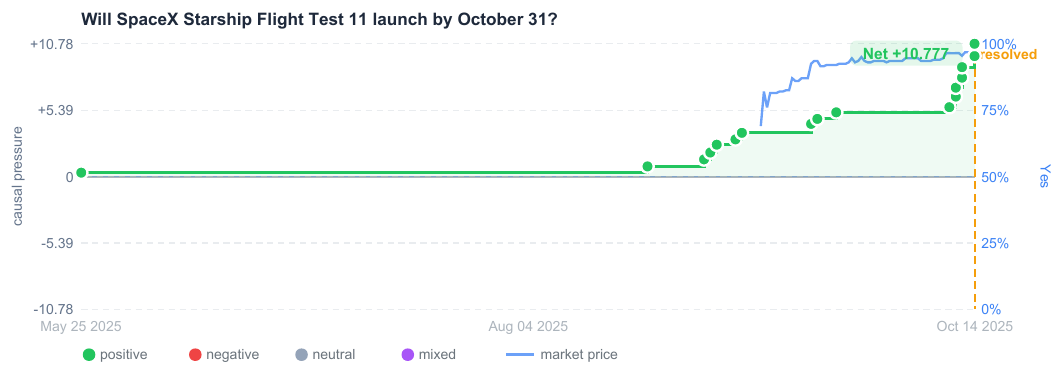}
\centerline{\small\textbf{(a)} Starship launch by October 31 (Yes)}
\end{minipage}

\vspace{1em}

\begin{minipage}{0.78\linewidth}
\centering
\includegraphics[width=\linewidth,height=0.20\textheight,keepaspectratio]{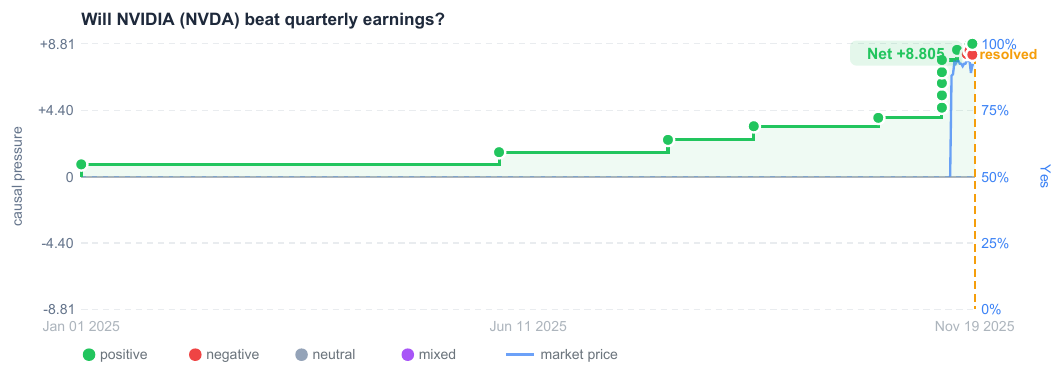}
\centerline{\small\textbf{(b)} NVIDIA quarterly earnings beat (Yes)}
\end{minipage}
\caption{Additional causal pressure charts. \textbf{(a)} Starship is a monotone positive-pressure case: all 17 hindsight events support the resolved outcome and market price rises with the pressure curve. \textbf{(b)} NVIDIA is a compressed-window case: the market was already high before most reference events arrived, illustrating why \proposedname\ evaluates event-level reasoning separately from final forecast accuracy.}
\label{fig:pressure_positive_examples}
\end{figure*}

\paragraph{Examples.}
Figure~\ref{fig:pressure_chatgpt} in the main paper shows a negative-pressure case. Figure~\ref{fig:pressure_positive_examples} adds two positive-outcome examples: Starship is a monotone case where event pressure and market price rise together, while NVIDIA is a compressed-window case where the market was already high before most reference events arrived.

%% file: sections/appendix/E_contamination.tex
\section{Knowledge-Cutoff Filtering and Parametric Leakage}
\label{app:contamination}

The knowledge-cutoff contamination filter excludes model-question pairs whose estimated start time precedes the model's recorded training cutoff. This is intentionally conservative: a cutoff date is only a coarse proxy for what appears in a model's parameters, but if a question was already open before that date, resolution-relevant information may have appeared in the training corpus. The filter is applied per model-question pair, not per question, because models have different cutoff dates.

This appendix examines the effect of the filter in the Vanilla LLM condition, where models receive only the question and cannot retrieve evidence or use tools. Vanilla performance is therefore the cleanest diagnostic for parametric leakage: any improvement on pre-cutoff questions must come from internal knowledge or generalization rather than benchmark evidence access. Table~\ref{tab:contamination_vanilla} compares unfiltered and filtered accuracy by model.

\begin{table*}[t]
\centering
\small
\setlength{\tabcolsep}{5pt}
\begin{tabular}{lcrrrrrr}
\toprule
\textbf{Model} & \textbf{Cutoff} & \textbf{All N} & \textbf{All Acc.} & \textbf{Filt.\ N} & \textbf{Filt.\ Acc.} & \textbf{Excl.} & \textbf{$\Delta$Acc.} \\
\midrule
Gemini 3 Pro    & 2025-01 & 120 & 67.5 & 108 & \textbf{66.7} & 12 & $-$0.8 \\
Gemini 3 Flash  & 2025-01 & 120 & 67.5 & 108 & 65.7 & 12 & $-$1.8 \\
DeepSeek V4 Pro & 2025-05 & 120 & 65.0 &  68 & 54.4 & 52 & $-$10.6 \\
DeepSeek V4 Flash & 2025-05 & 120 & 60.0 &  68 & 52.9 & 52 & $-$7.1 \\
GPT-5.4         & 2025-08 & 120 & \textbf{69.2} &  55 & 52.7 & 65 & $-$16.4 \\
GPT-4o          & 2023-10 & 120 & 53.3 & 118 & 53.4 &  2 & $+$0.1 \\
Qwen3.5 397B    & Unknown$^\dagger$ & 120 & 66.7 & 120 & 66.7 & 0 & $+$0.0 \\
\bottomrule
\end{tabular}
\caption{Vanilla LLM condition: unfiltered vs.\ contamination-filtered accuracy by model. \textit{Cutoff} is the model's recorded knowledge cutoff (year-month); \textit{Excl.}\ is the number of model-question pairs removed because the question start date precedes that cutoff. GPT-5.4's August 2025 cutoff overlaps substantially with the question pool, causing 54\% of pairs to be excluded and a 16.4-point drop. DeepSeek V4's May 2025 cutoff excludes 52 pairs, while Gemini's January 2025 cutoff excludes 12 pairs. GPT-4o (gpt-4o-2024-11-20) has an October 2023 cutoff; only 2 of 120 questions have start dates before that cutoff, leaving 118 clean pairs. Qwen3.5 397B has no official disclosed cutoff; $^\dagger$we therefore report it as an unfiltered diagnostic reference.}
\label{tab:contamination_vanilla}
\end{table*}

The filter has asymmetric effects across models. Gemini's January 2025 cutoff removes only 12 of 120 Vanilla pairs, and filtered accuracy changes by less than two points. DeepSeek V4's May 2025 cutoff removes 52 pairs, reducing Vanilla accuracy by 7.1 points for Flash and 10.6 points for Pro. GPT-4o's October 2023 cutoff removes only 2 pairs, so its filtered and unfiltered Vanilla scores are nearly identical. By contrast, GPT-5.4's August 2025 cutoff overlaps heavily with the benchmark's late-2025/early-2026 resolution window: 65 pairs are excluded and Vanilla accuracy falls from 69.2\% to 52.7\%. This is the clearest evidence that unfiltered knowledge-only evaluation can substantially overstate forecasting ability for models with recent training cutoffs.

Figure~\ref{fig:vanilla_time_performance_main} in the main text provides the corresponding temporal view, plotting accuracy against each question's estimated start time rather than its resolution date. The rolling curves should not be read as precise estimates for any single month because the evaluation set is small after stratification and the window is smoothed.

The diagnostic supports three qualitative conclusions. First, model comparisons are most reliable to the right of each model's cutoff line, where Vanilla performance cannot be attributed to post-start training exposure. Second, recent-cutoff models can look artificially strong in unfiltered Vanilla evaluation because many benchmark questions were already underway during their training horizon. Third, the post-cutoff performance drop shows why outcome accuracy alone is not enough: without the Temporal Gateway and evidence/reasoning metrics, a model's apparent forecasting ability may partly reflect stale or memorized knowledge rather than genuine prospective reasoning. For this reason, all main cross-model results report the contamination-filtered setting; unfiltered results are retained only as diagnostics.

%% file: sections/appendix/F_evaluation_metrics.tex
\section{Evaluation Metric Definitions}
\label{app:metrics}

This appendix gives the full formal definitions of the three-axis evaluation metrics introduced in Section~\ref{sec:benchmark}. Each axis maps to one or more scalar metrics; Table~\ref{tab:permodel_results} reports all three axes jointly.

\subsection{Outcome Quality}
\label{app:metrics:outcome}

Outcome quality measures whether the agent's probabilistic forecast is correct. Forecasts are stored as a predicted answer plus a confidence value. For binary questions, let $y_i \in \{0,1\}$ be the resolved outcome and let $p_i \in [0,1]$ be the probability assigned to the positive answer. For multiple-choice questions, the predicted option is the option with the highest reported confidence. Quantity and timeframe questions are included in accuracy, but Brier and log scores are reported only where a categorical probability score is defined.

\paragraph{Accuracy.}
\begin{equation}
\mathrm{Acc} = \frac{1}{N}\sum_{i=1}^{N} \mathbf{1}[\mathrm{correct}_i],
\end{equation}
where $\mathrm{correct}_i$ indicates that the submitted answer matches the resolved answer under the type-specific rules below. For binary questions, the hard prediction is $\hat{y}_i = \mathbf{1}[p_i \geq 0.5]$. For multiple-choice questions, $\hat{y}_i$ is the answer option with the highest assigned confidence.

For quantity questions, a point prediction is counted correct if it falls within 10\% of the resolved numeric value; interval predictions are counted correct if the resolved value lies inside the predicted interval. For timeframe questions, accuracy requires an exact normalized match to the resolved timeframe. When binary or multiple-choice predictions are phrased differently from the stored option string, an LLM judge is used only as a semantic-equivalence fallback.

\paragraph{Brier Score.}
\begin{equation}
\mathrm{BS} = \frac{1}{|\mathcal{C}|}\sum_{i \in \mathcal{C}}(p_i - y_i)^2,
\end{equation}
where $\mathcal{C}$ is the set of binary and multiple-choice forecasts for which a categorical confidence score is defined. Lower is better. Brier score penalises both direction errors and overconfidence.

For multiple-choice questions, we use the confidence assigned to the selected option: if the selected option is correct, the score is $(1-c_i)^2$; otherwise it is $c_i^2$. Because the current interface records a selected answer and confidence rather than a full probability vector over all options, this is a reduced multiclass scoring rule. Brier score is omitted for quantity and timeframe questions.

\paragraph{Log Score.}
\begin{equation}
\mathrm{Log} = \frac{1}{|\mathcal{C}|}\sum_{i \in \mathcal{C}}\log p_i^{(y_i)},
\end{equation}
where $\mathcal{C}$ is again the set of binary and multiple-choice forecasts with defined categorical confidence scores, and $p_i^{(y_i)}$ is the probability assigned to the resolved answer. Higher is better (less negative). Log score strictly rewards calibrated confidence and penalises near-zero probabilities on correct answers.

For multiple-choice questions, $p_i^{(y_i)}$ is the confidence assigned to the selected option if it is correct and the complement probability otherwise. As with Brier score, log score is omitted for quantity and timeframe questions. Condition- and model-level Brier/log averages are computed over forecasts for which the corresponding score is defined.

\subsection{Evidence Quality}
\label{app:metrics:evidence}

Evidence quality measures whether the agent cited or accessed sources that are causally relevant to the question outcome. Let $C_i$ be the set of sources accessed or cited by the agent, and $C_i^H$ the reference evidence set constructed by the hindsight pipeline.

\paragraph{Source Precision.}
\begin{equation}
\mathrm{SrcP} = \frac{1}{N}\sum_{i=1}^{N} \frac{|C_i \cap C_i^H|}{|C_i|},
\end{equation}
with $\mathrm{SrcP}_i = 0$ when $|C_i| = 0$. Precision is reported rather than recall because the agent's source budget is constrained and the hindsight set is exhaustive; we care whether the sources the agent chose were relevant, not whether it found every relevant source.

Source overlap is determined by exact article identifier match. Conditions that produce no retrieval (Vanilla LLM, Causal Simulation) receive no SrcP score; Real-Time is excluded because its sources lack comparable article provenance.

\begin{table*}[t]
\centering
\scriptsize
\setlength{\tabcolsep}{3.5pt}
\begin{tabular}{lc rrr rrr rrr}
\toprule
& & \multicolumn{3}{c}{\textit{Vanilla LLM}} & \multicolumn{3}{c}{\textit{Causal Sim.}} & \multicolumn{3}{c}{\textit{Search-Enabled}} \\
\cmidrule(lr){3-5}\cmidrule(lr){6-8}\cmidrule(lr){9-11}
\textbf{Model} & \textbf{N} & Acc & Br & Log & Acc & Br & Log & Acc & Br & Log \\
\midrule
Gemini 3 Pro      & 108 & \textbf{66.7} & \textbf{.210} & \textbf{$-$.661} & 59.3 & .241 & $-$.761 & \textbf{74.1} & .183 & $-$.611 \\
Gemini 3 Flash    & 108 & 65.7 & .244 & $-$.926 & \textbf{64.8} & \textbf{.214} & \textbf{$-$.639} & 72.2 & .203 & $-$.671 \\
DeepSeek V4 Pro   &  68 & 54.4 & .271 & $-$.770 & 57.4 & .251 & $-$.718 & 67.6 & .200 & $-$.616 \\
DeepSeek V4 Flash &  68 & 52.9 & .244 & $-$.705 & 57.4 & .263 & $-$.727 & 73.5 & \textbf{.146} & \textbf{$-$.442} \\
GPT-5.4           &  55 & 52.7 & .240 & $-$.679 & 54.5 & .235 & $-$.665 & 67.3 & .181 & $-$.543 \\
GPT-4o            & 118 & 53.4 & .258 & $-$.725 & 46.6 & .299 & $-$.806 & 59.3 & .273 & $-$1.153 \\
\midrule
\multicolumn{2}{l}{\textit{Qwen3.5 397B (diag., unfilt.)}} & 66.7 & .199 & $-$.582 & 71.7 & .191 & $-$.570 & 72.5 & .184 & $-$.567 \\
\midrule\midrule
& & \multicolumn{3}{c}{\textit{SE Graph}} & \multicolumn{3}{c}{\textit{Near-Resolution}} & \multicolumn{3}{c}{\textit{Real-Time}} \\
\cmidrule(lr){3-5}\cmidrule(lr){6-8}\cmidrule(lr){9-11}
\textbf{Model} & \textbf{N} & Acc & Br & Log & Acc & Br & Log & Acc & Br & Log \\
\midrule
Gemini 3 Pro      & 108 & \textbf{73.1} & .192 & $-$.685 & \textbf{79.6} & \textbf{.162} & $-$.671 & 91.7 & .077 & $-$.702 \\
Gemini 3 Flash    & 108 & 64.8 & .242 & $-$.934 & 76.9 & .182 & $-$.803 & 88.0 & .115 & $-$1.774 \\
DeepSeek V4 Pro   &  68 & 66.2 & .209 & $-$.632 & 76.5 & .174 & $-$.638 & \textbf{95.6} & \textbf{.028} & \textbf{$-$.107} \\
DeepSeek V4 Flash &  68 & 69.1 & .182 & $-$.542 & 77.9 & .171 & \textbf{$-$.590} & \textbf{95.6} & .030 & $-$.137 \\
GPT-5.4           &  55 & 58.2 & \textbf{.182} & \textbf{$-$.528} & 70.9 & .185 & $-$.602 & 81.8 & .138 & $-$.539 \\
GPT-4o            & 118 & 55.1 & .283 & $-$.969 & 66.9 & .238 & $-$.718 & 82.2 & .159 & $-$.736 \\
\midrule
\multicolumn{2}{l}{\textit{Qwen3.5 397B (diag., unfilt.)}} & 69.2 & .202 & $-$.605 & 73.3 & .180 & $-$.563 & 89.2 & .087 & $-$.315 \\
\bottomrule
\end{tabular}
\caption{Per-model results (contamination-filtered) across all six conditions (see Appendix~\ref{app:model_results}). \textit{Top}: knowledge-only and search-enabled conditions. \textit{Bottom}: graph-enabled and real-time conditions. Acc\,=\,accuracy (\%); Br\,=\,Brier; Log\,=\,log score. Bold indicates the best contamination-filtered model per condition and metric (higher is better except Br); Qwen3.5 397B is unfiltered and excluded from bolding. GPT-5.4 contributes 55 pairs (65 excluded); DeepSeek V4 models contribute 68 pairs (52 excluded).}
\label{tab:model_results_a}
\label{tab:model_results_b}
\end{table*}

\subsection{Reasoning Quality}
\label{app:metrics:reasoning}

Reasoning quality measures whether the agent's causal forecast graph recovers the events that proved most important to the outcome. Let $G_i^F = (V_i^F, E_i^F)$ be the agent's forecast graph and $G_i^H = (V_i^H, E_i^H)$ the hindsight reference graph.

\paragraph{Key Events.}
The reference key-event set $K_i \subseteq V_i^H$ consists of the top-$K$ hindsight events ranked by impact score $s_j = \mathrm{magnitude}_j \times \mathrm{confidence}_j$, where magnitude and confidence are assigned by the hindsight pipeline. We use $K = 5$ in all reported experiments.

\paragraph{Event Matching.}
A forecast event $v \in V_i^F$ matches a hindsight event $u \in V_i^H$ if two conditions hold simultaneously:
\begin{enumerate}
  \item \textbf{Text similarity}: the hybrid BM25+lexical score $\mathrm{sim}(v, u) \geq \theta$, where $\theta = 0.45$ and the score is $0.6 \times \mathrm{BM25\text{-}RS}(v,u) + 0.4 \times \mathrm{TokF1}(v,u)$. BM25-RS is BM25 reverse-scored to $[0,1]$; TokF1 is token-level F1 over unigrams.
  \item \textbf{Date proximity}: $|d(v) - d(u)| \leq \delta$ days, where $\delta = 7$ in all reported experiments.
\end{enumerate}
Each hindsight event is matched to at most one forecast event (greedy assignment by similarity score).

\paragraph{Key-Event Recall.}
\begin{equation}
\mathrm{KER} =
\frac{1}{N}\sum_{i=1}^{N}
\frac{1}{|K_i|}
\sum_{u \in K_i}
\mathbf{1}[\exists\, v \in V_i^F: M(v,u)] .
\end{equation}
Here \(M(v,u)\) denotes a successful textual and temporal match. KER measures what fraction of the most causally important hindsight events appear in the agent's forecast graph. Conditions that produce no graph (Vanilla LLM, Search-Enabled) receive no KER score.

\paragraph{Semantic Matching Robustness.}
The primary metric uses the hybrid lexical matcher because it is reproducible without model weights. Appendix~\ref{app:semantic_robustness} reports a parallel evaluation using sentence-transformer cosine similarity (\texttt{all-MiniLM-L6-v2}, threshold 0.55), confirming that all condition-level rankings are preserved under the alternative matcher, with scores uniformly 1--5\,pp higher due to paraphrase tolerance.

%% file: sections/appendix/G_model_results.tex

\section{Detailed Model Results}
\label{app:model_results}

Tables~\ref{tab:model_results_a} and~\ref{tab:model_results_b} report per-model contamination-filtered Brier score and log score across all six conditions, complementing the accuracy and reasoning metrics in Table~\ref{tab:permodel_results} of the main text. Table~\ref{tab:model_results_ker} gives the per-model reasoning and evidence quality breakdown. Qwen3.5 397B is included as an unfiltered diagnostic row; see Appendix~\ref{app:contamination} for contamination filter details.

\subsection{Outcome Quality}
\label{app:outcome_quality}

Table~\ref{tab:model_results_a} (in Appendix~\ref{app:metrics}) reports Brier and log scores across all six conditions. Higher accuracy and log score are better; lower Brier score is better.

\subsection{Reasoning and Evidence Quality}

Table~\ref{tab:model_results_ker} reports key-event recall (KER) and source precision (SrcP) per model. KER is only defined for graph-enabled conditions; SrcP is only defined for retrieval-enabled conditions with article provenance.

\begin{table*}[tp]
\centering
\scriptsize
\setlength{\tabcolsep}{4pt}
\begin{tabular}{l rr r rr rr rr}
\toprule
& \multicolumn{2}{c}{\textit{Causal Sim.}} & \textit{Search-En.} & \multicolumn{2}{c}{\textit{SE Graph}} & \multicolumn{2}{c}{\textit{Near-Res.}} & \multicolumn{2}{c}{\textit{Real-Time}} \\
\cmidrule(lr){2-3}\cmidrule(lr){4-4}\cmidrule(lr){5-6}\cmidrule(lr){7-8}\cmidrule(lr){9-10}
\textbf{Model} & KER & SrcP & SrcP & KER & SrcP & KER & SrcP & KER & SrcP \\
\midrule
Gemini 3 Pro      & 5.2 & --   & \textbf{70.4} & \textbf{12.2} & \textbf{66.6} & 13.3 & \textbf{71.6} & 10.4 & -- \\
Gemini 3 Flash    & 5.2 & --   & 58.8 &  6.4 & 52.5 & 11.2 & 64.8 &  9.8 & -- \\
DeepSeek V4 Pro   & 5.7 & --   & 65.2 & \textbf{13.7} & 58.7 & \textbf{20.4} & 71.4 & 12.0 & -- \\
DeepSeek V4 Flash & 4.7 & --   & 65.2 & 12.0 & 57.4 & 18.1 & 66.6 & \textbf{15.0} & -- \\
GPT-5.4           & \textbf{6.3} & --   & 61.2 &  9.5 & 54.3 & 14.9 & 68.1 &  9.8 & -- \\
GPT-4o            & 3.4 & --   & 59.7 &  7.2 & 56.0 & 12.6 & 70.4 &  8.2 & -- \\
\bottomrule
\end{tabular}
\caption{Per-model reasoning and evidence quality (contamination-filtered). KER\,=\,key-event recall (\%, 7-day date gate); SrcP\,=\,source precision (\%). Bold indicates the best contamination-filtered model within each condition and metric. Causal Simulation has no retrieval so SrcP is not applicable; Real-Time lacks article provenance so SrcP is omitted. Vanilla LLM and Search-Enabled have no graph so KER is not applicable.}
\label{tab:model_results_ker}
\end{table*}

\subsection{Per-Model Conditional Quality}
\label{app:conditional_quality}

Table~\ref{tab:conditional_quality} in the main text reports the aggregate correctness-conditioned split. Table~\ref{tab:conditional_quality_model} gives the same split by model for the Search-Enabled Graph condition, where both graph and source metrics are defined. These rows should be read as exploratory diagnostics rather than model rankings, because the correct/incorrect splits are small after contamination filtering, especially for recent-cutoff models.

\begin{table*}[tp]
\centering
\scriptsize
\begin{tabular}{lrrrrrrrr}
\toprule
& \multicolumn{3}{c}{\textit{Correct forecasts}} & \multicolumn{3}{c}{\textit{Incorrect forecasts}} & \multicolumn{2}{c}{\textit{Gap}} \\
\cmidrule(lr){2-4}\cmidrule(lr){5-7}\cmidrule(lr){8-9}
\textbf{Model} & \textbf{N} & \textbf{KER} & \textbf{SrcP} & \textbf{N} & \textbf{KER} & \textbf{SrcP} & $\Delta$\textbf{KER} & $\Delta$\textbf{SrcP} \\
\midrule
Gemini 3 Flash    & 70 &  7.2 & 55.5 & 38 &  4.9 & 46.7 &  2.4 &  8.8 \\
Gemini 3 Pro      & 79 & 12.9 & 74.4 & 29 & 10.3 & 45.3 &  2.6 & 29.1 \\
DeepSeek V4 Flash & 47 & 13.2 & 56.4 & 21 &  9.3 & 60.0 &  3.9 & $-$3.6 \\
DeepSeek V4 Pro   & 45 & 14.2 & 61.0 & 23 & 12.6 & 53.6 &  1.6 &  7.4 \\
GPT-4o            & 65 &  8.3 & 64.6 & 53 &  5.9 & 44.8 &  2.4 & 19.8 \\
GPT-5.4           & 32 & 10.0 & 60.0 & 23 &  8.9 & 46.1 &  1.1 & 13.9 \\
\bottomrule
\end{tabular}
\caption{Per-model conditional quality for the Search-Enabled Graph condition (contamination-filtered). KER and SrcP are percentages. Gaps are correct minus incorrect, so positive values mean the metric is higher among correct forecasts. These splits are diagnostic only: each row is further divided by correctness, and DeepSeek V4/GPT-5.4 have smaller filtered sample sizes because of later knowledge cutoffs.}
\label{tab:conditional_quality_model}
\end{table*}

\subsubsection{Forecast Graph Alignment with Market Moves}

For Polymarket questions, forecast-time graph quality can also be evaluated against market movement. Unlike the hindsight reference graph in Appendix~\ref{app:pressure_charts}, forecast graphs do not provide a common signed pressure curve against the resolved outcome; they contain model-generated events and causal links whose node set varies by run. We therefore use market prices to define an external subset of important reference events, then ask whether each forecast graph recovers those events. Market-signal recall is the fraction of market-aligned hindsight events recovered by the forecast graph, where a hindsight event is market-aligned when its signed impact agrees with a nearby price move. Market-signal precision is the fraction of forecast graph events that match such market-aligned hindsight events. We additionally report a stricter turning-point variant that only counts market-aligned hindsight events occurring near detected market turning points.
Table~\ref{tab:forecast_market_signal_recall} reports this market-aligned recovery split by model and graph-enabled condition.

\begin{table*}[tp]
\centering
\scriptsize
\begin{tabular}{llrrrr}
\toprule
\textbf{Condition} & \textbf{Model} & \textbf{Mkt R} & \textbf{Mkt P} & \textbf{TP R} & \textbf{TP P} \\
\midrule
Causal Sim. & Gemini 3 Flash & 2.0\% & 2.6\% & 2.1\% & 2.3\% \\
& Gemini 3 Pro & 0.9\% & 0.7\% & 1.2\% & 0.9\% \\
& DeepSeek V4 Flash & 2.6\% & 1.7\% & 1.1\% & 0.7\% \\
& DeepSeek V4 Pro & 1.6\% & 1.1\% & 0.9\% & 0.4\% \\
& GPT-4o & 0.3\% & 0.2\% & 0.0\% & 0.0\% \\
& GPT-5.4 & \textbf{2.8\%} & \textbf{3.9\%} & \textbf{2.5\%} & \textbf{3.1\%} \\
\midrule
SE Graph & Gemini 3 Flash & 8.8\% & 5.7\% & 6.9\% & 4.2\% \\
& Gemini 3 Pro & 8.1\% & 6.5\% & 4.9\% & 4.6\% \\
& DeepSeek V4 Flash & \textbf{15.0\%} & 6.8\% & \textbf{11.1\%} & \textbf{4.9\%} \\
& DeepSeek V4 Pro & 14.4\% & \textbf{8.0\%} & 9.7\% & 4.0\% \\
& GPT-4o & 8.4\% & 3.1\% & 6.4\% & 2.2\% \\
& GPT-5.4 & 12.3\% & 6.1\% & 9.4\% & 4.7\% \\
\bottomrule
\end{tabular}
\caption{Per-model forecast graph recovery of market-aligned hindsight events. Mkt R/P use local before/after price movement around each hindsight event. TP R/P use the stricter subset requiring the hindsight event to fall within the 3-day market window of a detected turning point. Bold indicates the best model within each condition and metric.}
\label{tab:forecast_market_signal_recall}
\end{table*}

\subsubsection{Semantic Matching Robustness}
\label{app:semantic_robustness}

Our primary reasoning evaluation uses BM25+lexical hybrid matching (0.6$\times$BM25 reverse-scored + 0.4$\times$token-F1, threshold\,=\,0.45) because it is reproducible without model weights or API calls. To validate that this choice does not introduce systematic bias, we re-ran the full evaluation using sentence-transformer cosine similarity (\texttt{all-MiniLM-L6-v2}, threshold\,=\,0.55) under the same 7-day date gate and compared the results.
Table~\ref{tab:semantic_robustness} shows that semantic matching raises absolute scores while preserving the condition ordering.

\begin{table}[t]
\centering
\small
\resizebox{\columnwidth}{!}{%
\begin{tabular}{llrrl}
\toprule
\textbf{Condition} & \textbf{Metric} & \textbf{Hybrid} & \textbf{Semantic} & \textbf{$\Delta$} \\
\midrule
Causal Sim.  & Event F1     & 13.1\% & 13.9\% & $+$0.8pp \\
             & KE Recall    &  4.9\% &  6.9\% & $+$2.0pp \\
             & KE Precision &  5.1\% &  7.5\% & $+$2.4pp \\
\midrule
SE Graph     & Event F1     & 17.8\% & 20.4\% & $+$2.6pp \\
             & KE Recall    &  9.8\% & 14.2\% & $+$4.4pp \\
             & KE Precision & 10.1\% & 14.2\% & $+$4.2pp \\
\midrule
Near-Res.    & Event F1     & 20.7\% & 24.5\% & $+$3.8pp \\
             & KE Recall    & 14.4\% & 19.6\% & $+$5.1pp \\
             & KE Precision & 14.6\% & 19.7\% & $+$5.1pp \\
\midrule
Real-Time    & Event F1     & 17.4\% & 20.9\% & $+$3.4pp \\
             & KE Recall    & 10.6\% & 14.6\% & $+$4.1pp \\
             & KE Precision & 12.9\% & 17.3\% & $+$4.4pp \\
\bottomrule
\end{tabular}}
\caption{Hybrid BM25+lexical vs.\ sentence-transformer semantic matching (\texttt{all-MiniLM-L6-v2}, threshold\,=\,0.55), both under a 7-day date gate. Semantic matching yields uniformly higher scores ($+$1--5\,pp), as it recognises paraphrases that lexical overlap misses; however, all relative rankings between conditions are preserved, confirming that the lexical evaluation is a valid lower bound for comparative analysis.}
\label{tab:semantic_robustness}
\end{table}

The robustness check shows that the lexical matcher is conservative rather than misleading. Semantic matching raises all reasoning scores by 1--5 percentage points, with the largest gains for key-event recall in Near-Resolution and SE Graph settings, where agents often describe the same event using different wording from the hindsight graph. This is expected: paraphrase-tolerant matching recovers semantically equivalent event mentions that fail token-level overlap.

Crucially, the condition ordering is unchanged. Near-Resolution remains the strongest graph-enabled setting, SE Graph remains above Causal Simulation, and Real-Time remains competitive but not directly comparable because it lacks the same simulated-date evidence constraint. We therefore report the hybrid lexical metric in the main tables as a reproducible lower bound, while using the semantic check to verify that the main conclusions do not depend on a brittle string-matching choice.

%% file: sections/appendix/H_frontend_inspection.tex
\begin{figure}[t]
\centering
\begin{minipage}{0.82\linewidth}
\centering
\includegraphics[width=\linewidth,height=0.34\textheight,keepaspectratio]{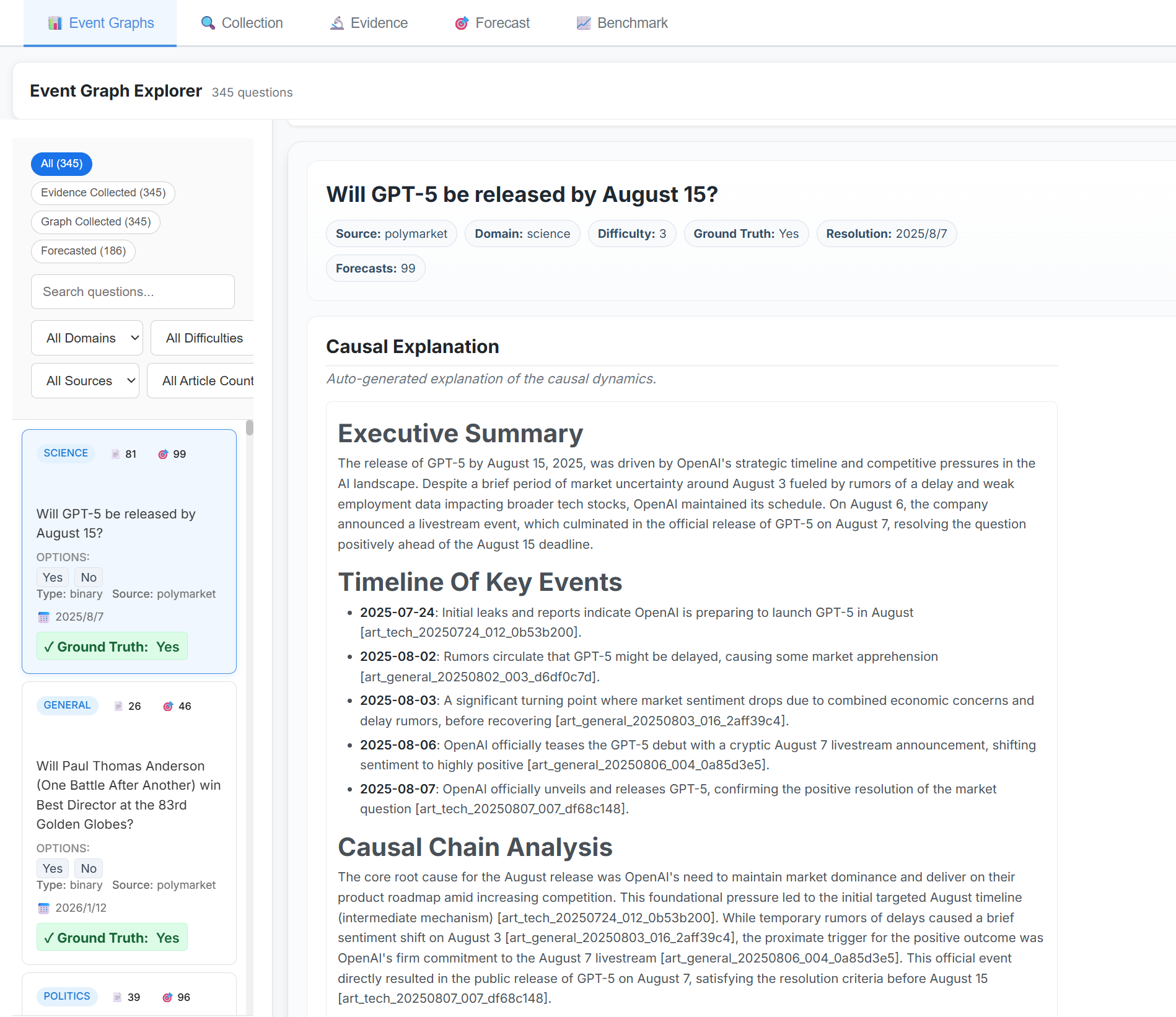}
\vspace{-0.5em}
\centerline{\small\textbf{(a)} Event-graph explorer for a resolved benchmark question}
\end{minipage}

\vspace{0.55em}

\begin{minipage}{0.82\linewidth}
\centering
\includegraphics[width=\linewidth,height=0.18\textheight,keepaspectratio]{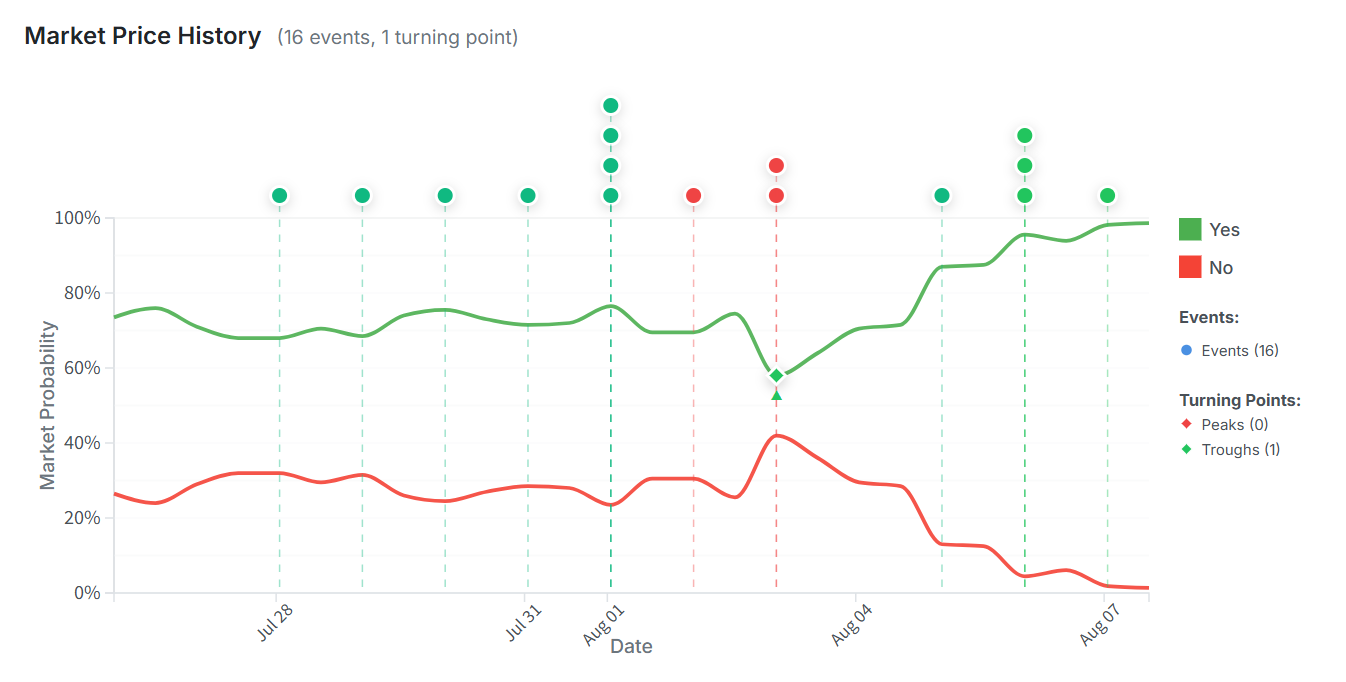}
\vspace{-0.5em}
\centerline{\small\textbf{(b)} Market-history view with extracted events and turning points}
\end{minipage}
\caption{Frontend inspection views for \proposedname. \textbf{(a)} The event-graph explorer exposes each resolved question with metadata, ground truth, forecast count, source-backed causal explanation, and dated key events, supporting manual audit of benchmark artifacts. \textbf{(b)} The market-history view overlays prediction-market prices with extracted hindsight events and turning points, supporting inspection of whether post-resolution event graphs align with external market signals. These interfaces are used for benchmark construction and review; forecast agents are evaluated through the controlled Temporal Gateway rather than through these UI views.}
\label{fig:frontend_inspection}
\end{figure}

\section{Frontend Inspection Views}
\label{app:frontend_inspection}

\proposedname\ includes a frontend for inspecting benchmark artifacts during construction, review, and analysis. These views are not shown to forecast agents during evaluation; agents interact with the benchmark through the controlled Temporal Gateway and the configured agent tools. The frontend instead supports auditability by exposing resolved questions, source-backed hindsight explanations, event timelines, and market-alignment checks in a human-readable form. Figure~\ref{fig:frontend_inspection} shows two representative inspection views.

%% file: sections/appendix/I_ai_use.tex
\section{Use of AI Assistance}
\label{app:ai_use}

AI assistants were used to support coding, data-processing checks, LaTeX editing, and writing revision during the preparation of this work. The authors reviewed, edited, and approved all final text, analyses, experimental results, and scientific claims.

%% file: sections/appendix/J_artifact_terms.tex
\section{Artifact License and Terms}
\label{app:artifact_terms}

The \proposedname\ code and derived benchmark artifacts are released for research use at \href{https://github.com/cyzus/worldreasoner/}{github.com/cyzus/worldreasoner}, subject to the licenses and terms of the original data sources. The released artifacts include generated forecasting questions, metadata, evaluation scripts, model outputs, and derived event-graph annotations. Source articles and prediction-market records may be represented through identifiers, metadata, timestamps, and derived summaries rather than redistributed wholesale when source terms restrict redistribution.

Users should comply with the applicable licenses of any underlying news, prediction-market, or model-provider data sources; the artifact is intended for research evaluation and should not be used as a deployed decision-support system for high-stakes forecasting.